\documentclass[11pt]{article}
\usepackage[numbers]{natbib}
\usepackage{macros/packages}
\usepackage{macros/editing-macros}
\usepackage{macros/formatting}
\usepackage{macros/statistics-macros}
\usepackage[utf8]{inputenc} 
\usepackage[T1]{fontenc}    
\usepackage{hyperref}       
\usepackage{url}            
\usepackage{booktabs}       
\usepackage{amsfonts}       
\usepackage{nicefrac}       
\usepackage{microtype}      
\usepackage{xcolor}         
\usepackage{wrapfig}
\usepackage{subcaption}
\usepackage{tikz}
\usetikzlibrary{%
  calc,
  fit,
  arrows,
  arrows.meta,
  positioning,
  decorations.pathreplacing,
  decorations.shapes,
}

\begin{document}

\abovedisplayskip=8pt plus0pt minus3pt
\belowdisplayskip=8pt plus0pt minus3pt


\begin{center}
  {\huge Exchangeable Sequence Models Quantify Uncertainty Over Latent Concepts} \\
  \vspace{.5cm} {\Large Naimeng Ye ~~~ Hongseok Namkoong} \\
  \vspace{.2cm}
  {\large Columbia University} \\
  \vspace{.2cm}
  \texttt{naimeng.ye@columbia.edu, namkoong@gsb.columbia.edu}
\end{center}


\begin{abstract}%
  Intelligent agents must be able to articulate its own uncertainty.  In this
work, we show that pre-trained sequence models are naturally capable of
probabilistic reasoning over exchangeable data points---forming informed
beliefs and sharpening them as it gathers more information. A sequence model
learns the relationship between observations, which differs from typical
Bayesian models that quantify uncertainty over latent parameters through
priors and likelihoods (e.g., topic models). Despite the apparent difference,
we illustrate how exchangeable sequence modeling provides a valid Bayesian
model by going back to De Finetti's classical \emph{predictive} view of
probabilistic reasoning: uncertainty comes from data that has not been
observed yet, rather than latent parameters.  From this perspective,
pre-training autoregressive models is equivalent to formulating informed
beliefs based on prior observations (``empirical Bayes''), and forward
generation is equivalent to simulating instantiations of an environment
(``posterior inference'').  In particular, exchangeable sequence models can
explicitly perform statistical inference; epistemic uncertainty over latent
environments is captured by variation in predicted future observations.
Formally, we show the sequence prediction loss controls the quality of
uncertainty quantification, and propose several approaches for encoding
exchangeability in sequence model architectures: data augmentation,
regularization, and causal masking.


\end{abstract}

\section{Introduction}
\label{section:introduction}

Intelligent systems must be able to utilize the information gathered so far to
gauge uncertainty on the underlying environment they are interacting with. For
example, given a sequence of questions and answers with a patient, an
intelligent model should be able to maintain an internal understanding of its
level of uncertainty on their mental health condition. Following standard
terminology in Bayesian statistics, we refer to uncertainty on the underlying
state as \emph{epistemic}. This uncertainty is fully reducible if the agent is able to observe a large, potentially infinite,  set of questions and answers with the patient.

Systems that can reason through epistemic uncertainty based on natural
language feedback has been a longstanding challenge. On the other hand,
autoregressive models pre-trained on massive web data exhibit striking
\emph{predictive} capabilities when conditioned on even a small number of
demonstrations~\cite{BrownEtAl20}.  ``In-context learning'' (ICL) has thus
emerged as a powerful learning paradigm where autoregressive generation
provides a versatile pattern recognition model without explicit training, even
on complex tasks like mathematical reasoning~\cite{WeiEtAl22, DongEtAl22}.
Several recent works study \emph{predictive uncertainty}---model confidence in
predictions---and provide interpretations of autoregressive probabilities as
Bayesian posterior predictive distributions~\cite{MullerHoArGrHu22,
  NguyenGr22, JeonLeLeVa24} and study their
calibratedness~\cite{KadavathEtAl22, TianEtAl23, LyuEtAl24}.  

In this work, we go beyond predictive uncertainty in a single inference step
and focus on a dimension of ICL that has received little attention.
\begin{itbox}
  When can pre-trained sequence models reason about its own level of epistemic
  uncertainty on latent environment (if any)?
\end{itbox}
We illustrate how insights that date back to~\citet{DeFinetti33} highlight
inferential capabilities of ICL, going beyond the predictive paradigms studied
in prior works~\cite{GargTsLiVa22,BaiChWaXiMe23}.  In the context of the
mental health example above, we are interested in the agent's uncertainty over
the patient's mental state (epistemic uncertainty over latent environment),
rather than the answer to the next question (predictive uncertainty).

\begin{figure}[t]
    \centering
    \includegraphics[width=\textwidth]{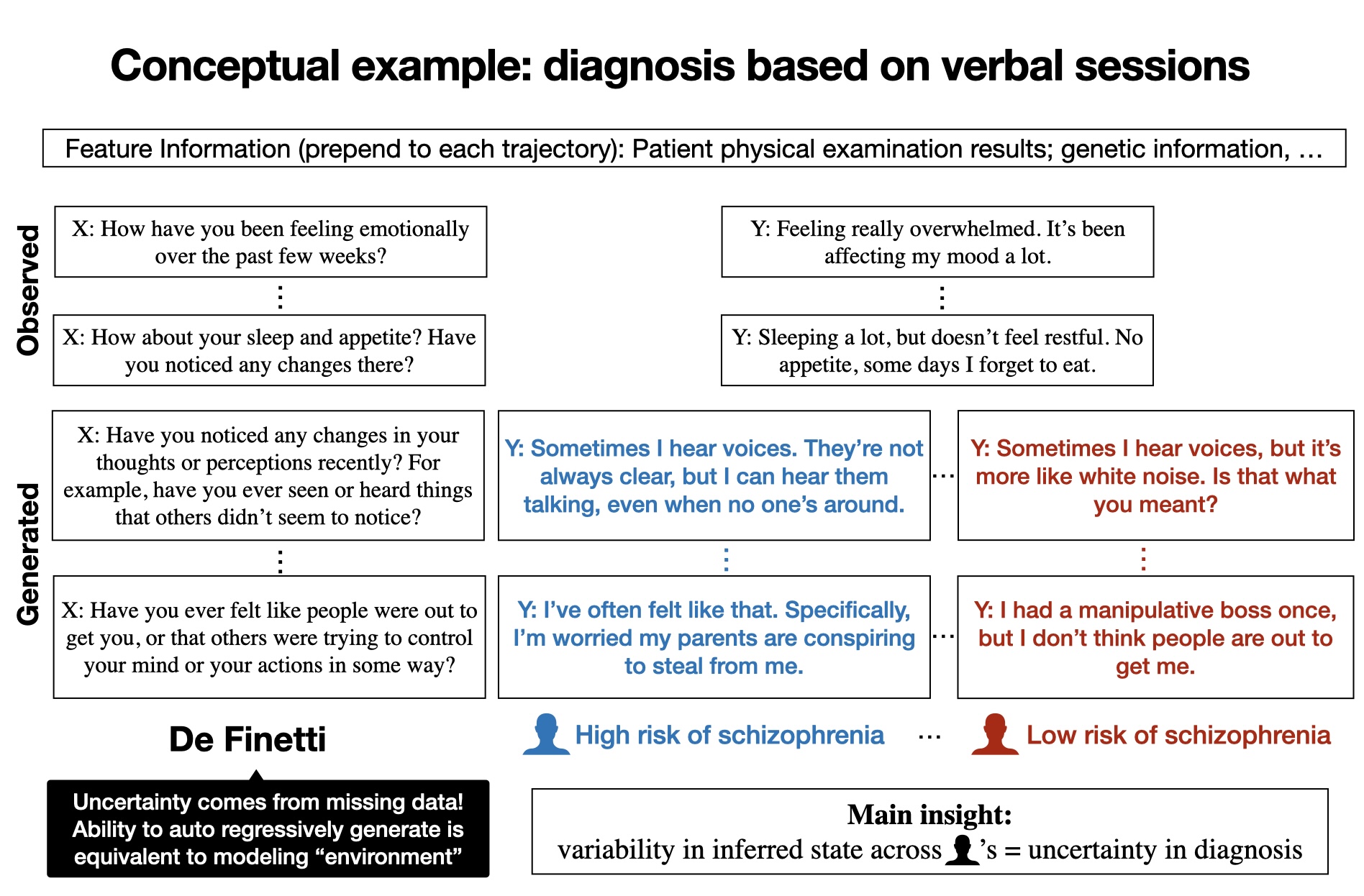}
    \caption{\citet{DeFinetti33}'s predictive view uncertainty in latent
      environment (mental state of the patient) as coming from future data
      (questions and answers). Building on this insight, we show the sequence
      prediction loss (perplexity) over exchangeable documents measures the
      quality of uncertainty quantification over latent environments. Thus,
      standard pre-training methods are in fact directly optimizing them
      through auto-differentiation and GPU parallelization. }
  \label{fig:definetti-overview}
\end{figure}

We begin by briefly reviewing competing views of Bayesian reasoning over
latent environments. The traditional Bayesian modeler posits a prior and
likelihood on latent variables that are fundamentally unobservable. This is
often challenging since even the notion of a latent mental state of a patient
is infinite-dimensional and ill-defined. On the other hand, De Finetti focuses
on modeling \emph{observables} rather than latents~\cite{DeFinetti33}. This
predictive view models uncertainty as coming from future data that has not
been observed yet (see Figure~\ref{fig:definetti-overview}).

\begin{figure}[t]
  \centering
  \begin{subfigure}[b]{0.48\textwidth}
    \centering
    \includegraphics[width=\textwidth, height=0.27\textheight]{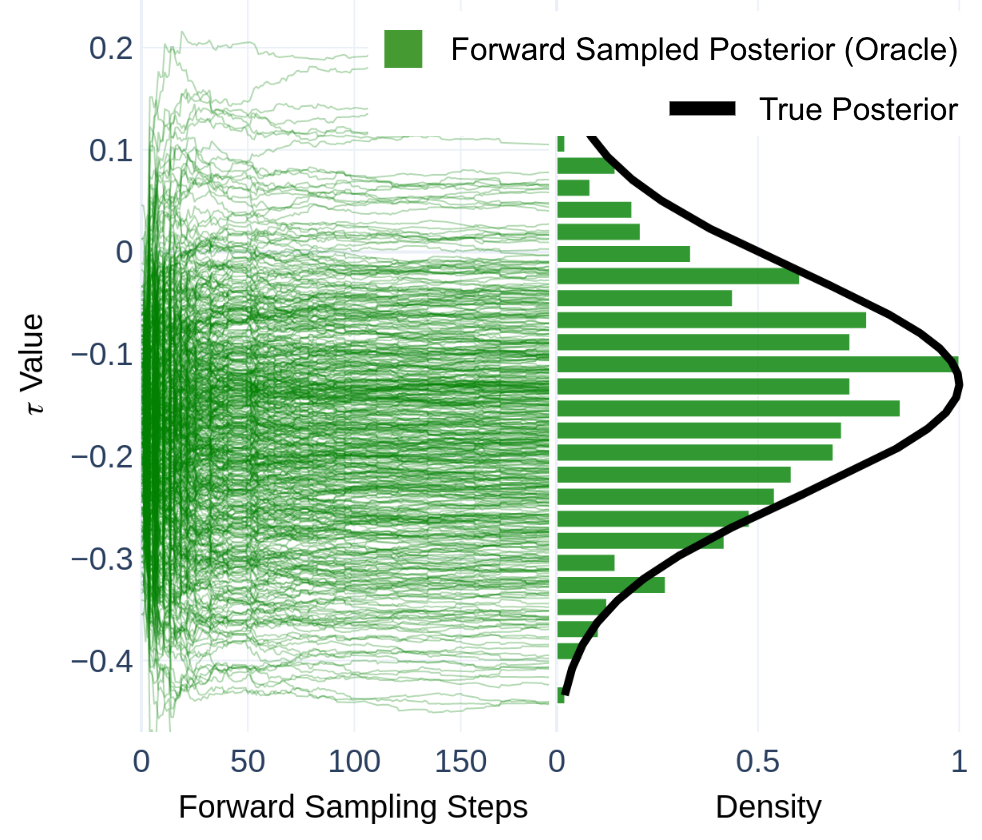}
    \caption{}\label{fig:icl-inference}
  \end{subfigure}
  \hfill
  \begin{subfigure}[b]{0.48\textwidth}
    \centering
    \includegraphics[width=\textwidth]{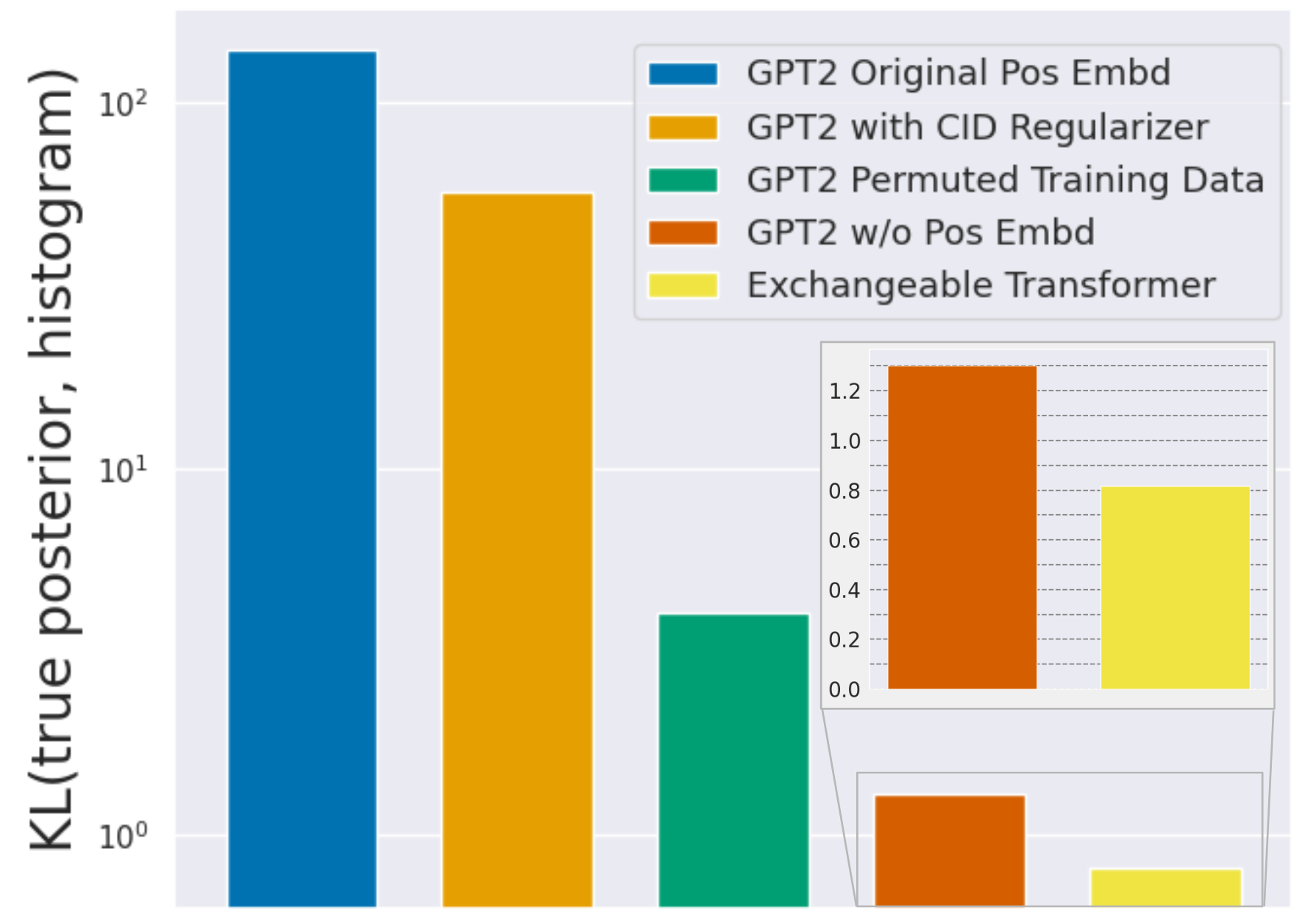}
    \caption{}\label{fig:diff}
  \end{subfigure}
  \caption{(a) Given an observed sequence (``prompt''), autoregressive
    generation provides inferential capabilities by computing a statistic over
    the generated trajectory. The panel on the left plots trajectories of
    forward generated outcomes; the panel on the right plots the histogram of
    the empirical mean of a trajectory. Under permutation invariance
    (exchangeability), this histogram is a valid approximation of the
    posterior distribution over the population mean. (b)~Autoregressive models
    provide approximate posterior draws via forward sampling. We plot the KL
    divergence between this approximate posterior of a latent parameter and the posterior produced by the oracle. Our
    experiments show that enforcing exchangeability via causal masking
    (Figure~\ref{fig:et}) provides large gains in inferential capabilities
    with 41x less parameters.}
  \label{fig:combined}
\end{figure}

Formally, De Finetti-Hewitt-Savage~\cite{DeFinetti33, DeFinetti37, HewittSa55}
showed that for a exchangeable sequence---whose joint distribution is
invariant to permutations---there is a latent random variable
(``environment'') that governs the data generation process: conditional on the
random variable, the inputs are i.i.d.. Traditional Bayesian models use this
result as a justification for latent factor models: probabilistic topic models
posit a prior over latent topics, conditioned on which documents are generated
from a likelihood~\cite{BleiNgJo03}. Instead, we follow De Finetti's
predictive view and treat autoregressive generation of future data as a
simulated instantiation of an environment. Variations in the simulated future
data naturally capture uncertainty on the latent environment, allowing us to
perform posterior inference (see Figure~\ref{fig:combined}a for an
illustration).

Our main (tautological) observation is that instead of modeling latent
variables that are never observed, autoregressive models consider a sequence
prediction problem over \emph{observables}. If a sequence model satisfies
exchangeability---invariant under permutations---it defines a proper Bayesian
inference machine through autoregressive generation of sequences
(Section~\ref{section:de-finetti}).
\begin{itbox}
  Pre-trained sequence models are able to reason about uncertainty over latent
  parameters that govern a permutation invariant (exchangeable) set of
  documents.
\end{itbox}


Viewing ICL as a Bayesian statistician, we expand previously proposed
downstream ICL tasks to include those that require comprehending uncertainty.
First, we consider \emph{length generalization} where we wish to achieve
robust predictive performance over sequences longer than that seen during
pre-training (Sections~\ref{section:length-gen}). Second, we consider
\emph{statistical inference} where we wish to develop valid confidence
intervals on parameters governing data generation
(Section~\ref{section:inference}).  We show that the sequence prediction loss
directly controls performance on these downstream tasks. Our theory explains
permutation invariance in autoregressive models allows robust ICL performance
on long sequences and protection against distribution shift from pre-training
to ICL.

From a modeling perspective, transformer-based sequence models with positional
embeddings are not naturally permutation invariant over \emph{documents}
(self-attention is permutation invariant over tokens without positional
embeddings~\cite{VaswaniEtAl17}). Empirically, we explore several different
methods for instilling permutation invariance in a transformer; in
Figure~\ref{fig:combined}b, we compare data augmentation (green), loss-based
regularizers (orange), and causal masking schemes (orange). We analyze their
impact on the above two tasks where comprehending epistemic uncertainty is
critical, and find the direct modification to the causal masks is the most
effective. Since the length of each document is variable in practice, our
preliminary numerical study highlights future research directions for
architecture design.

The main contribution of this work is conceptual rather than practical.
Following prior work on ICL~\cite{GargTsLiVa22}, we use contrived yet
principled examples to articulate our insight. Beyond these simple examples,
we hope our perspective can open up new applications of ICL with uncertainty
quantification as a central focus. For example, imagine a psychiatric risk
assessment system that informs the next course of action (e.g., prioritization
of cases for clinicians) based on a natural language-based patient survey.
Our results show that based on the observed response to the survey
questionnaire, autoregressively simulating answers to unobserved questions
provides proper Bayesian inference on the patient's mental state. Notably, a
key feature of our framework is that the model can develop sharper beliefs as
it observes more answers from the patient, which in turn can lead to granular
subsequent interventions.


While the connection between ICL and Bayesian reasoning is folklore at this
point (e.g.,~\cite{XieRaLiMa21, JeonLeLeVa24}), we substantiate it by exactly
characterizing ICL as \emph{explicit} Bayesian inference. Our observation is
tautological, and aside from exchangeability, it does not rely on particular
architectures nor elaborate data generation models considered in prior
work~\cite{XieRaLiMa21}.  We emphasize we are not the first to make this
observation: it is due to De Finetti~\cite{DeFinetti33, DeFinetti37,
  CifarelliRe96, DeFinetti17} and decidedly classical! Our main contribution
is of an expository nature, crystallizing and contextualizing this insight in
the modern context of pre-training and ICL.  We connect several disparate
works on Bayesian statistics~\cite{DeFinetti33, Roberts65, Ericson69, Dawid92,
  DawidVo97, BertiReRi98, FortiniLaRe00, FortiniPe14, HahnMaWa18,
  BertiDrPrRi21, BertiDrLePrRi22, FongHoWa23}, meta
learning~\cite{MullerHoArGrHu22, NguyenGr22}, and Bayesian deep
learning~\cite{GarneloRoMaRaSaShTeReEs18}, and demonstrate ICL can go beyond
predictions and quantify epistemic uncertainty through forward generation.
Our work contributes to the nascent literature that builds formal models of
ICL using hidden Markov models~\cite{XieRaLiMa21}, statistical learning with
stability conditions~\cite{LiIlPaOy23}, or gradient descent-based
algorithms~\cite{DaiSuDoHaMaSuWe23, BaiChWaXiMe23, VonOswaldEtAl23,
  AkyurekScAnMaZh23}.

Several concurrent works study De Finetti's predictive view of
probabilistic reasoning.  \citet{ZhangMcSuZhGr23} illustrates this connection
using topic models and demonstrate how LLMs can recover latent topic
distributions. \citet{ZhangCaNaRu24} proposes to train sequence models to
solve meta-bandit problems where active exploration is necessary.
\citet{FalckWaHo24} argues for the necessity of a key coherence
property---autoregressive probabilities form a martingale---which extends
exchangeability.  Compared to these works, we illustrate De Finetti's insight
in the context of ICL, and expand the scope of ICL to tasks that require
uncertainty quantification. Computationally, we propose and study different
forms of inductive bias for exchangeability in sequence modeling.


\section{Preliminaries}
\label{section:background}

In this section, we set up basic notation required to contrast the two
approaches to Bayesian modeling: the traditional Bayesian view that models
latent parameters, and De Finetti's predictive view that directly models
sequence of observables. We briefly review the former in this section, before
moving to the latter in the next.

Our subsequent results and algorithms can consider covariates/features for
each example, but we ignore them to simplify exposition and only mention the
generalization as a comment in theorem statements.

\subsection{Traditional Bayesian modeling}
\label{sec:two-approaches}

A traditional Bayesian statistician posits a latent parameter $\theta$ (e.g.,
a patient's high-dimensional mental state), a distribution over the latent,
$\pi_{\theta}$ (``prior''), and how $\theta$ governs data generation (e.g.,
patient interactions given mental state),
$\P(\Obs_{1:t} = \cdot \mid \theta) = P_\theta(\Obs_{1:t})$
(``likelihood''). A typical Bayesian model posits observables are
conditionally i.i.d., where the likelihood factorizes as
\begin{equation}
  \label{eqn:conditional iid}
  \P(\Obs_{1:\infty} = \obs_{1:\infty})
  = \int \prod_{t=1}^\infty \P(\Obs_t = \obs_t \mid \theta) \pi(d \theta)
\end{equation}
for some latent $\theta\sim \pi(\cdot)$ drawn from some prior.

We use the term \emph{epistemic} uncertainty to refer to the modeler's
reducible uncertainty and the term \emph{aleatoric} uncertainty to refer to
the inherent randomness in the data. Then, given observed data $\Obs_{1:s}$,
the posterior distribution $\pi(\theta = \cdot \mid \Obs_{1:s})$ measures the
epistemic uncertainty on the latent parameter. Usually, this posterior is
gradually reduced to the usual dirac measure (on the true latent) as more data
is gathered. The likelihoods represent the aleatoric uncertainty, since the
randomness is intrinsic to the data-generating process. To summarize, in
traditional Bayesian modeling:

\begin{align*}
  \text{Prior:} \quad & \pi(\theta) \\
  \text{Likelihood:} \quad & \P(\Obs_{1:t} = \cdot \mid \theta) = P_\theta(\Obs_{1:t}) \quad \quad \text{ aleatoric uncertainty}\\
  \text{Posterior:} \quad & \pi(\theta = \cdot \mid \Obs_{1:s}) \quad \quad \quad\quad \quad\qquad\text{epistemic uncertainty}\\
  \text{Posterior predictive:} \quad & \P(\Obs_{s+1:T} = \cdot \mid \Obs_{1:s}) = \int \P(\Obs_{s+1:T} = \cdot \mid \theta) \pi(\theta = \cdot \mid \Obs_{1:s}) d\theta.
\end{align*}

The main challenge with this modeling paradigm is the need to specify a model
over latent factors and argue for its validity despite its fundamentally
unobservable nature. While its philosophical standing is subject to debate,
a practical model validation metric
is to check whether the posited model on latent parameters explain observed
data well~\cite{GelmanCaStDuVeRu13}: for a posited model $\what{p}$, the
\emph{marginal likelihood} measures whether $\what{p}$ explains observed
sequences
\begin{equation*}
  \risk_n(\what{p}) \defeq \frac{1}{n} \sum_{i=1}^n \log \int \what{p} \left(
    \Obs_{1:T} \mid \theta \right) \pi(\theta) d\theta.
\end{equation*}
Tuning hyperparameters of the prior $\what{p}(\theta)$ based on this measure
is often called \emph{empirical Bayes}. Bayesian deep learning methods
subscribe to this view and propose model designs that aim to capture this
latent structure~\cite{Blundell15, OsbandAsCa18, WangYe20, JospinLaBoBuBe22}. 
Here, we use $\what{p}$ to denote the model's general form, which can implicitly generate prior, likelihood, and posterior distributions.

\subsection{De Finetti's theorem and the role of exchangeability}
\label{sec:de finetti's theorem}

De Finetti's characterization of an exchangeable sequence $\Obs_{1:\infty}$
(i.e., its distribution is permutation invariant) provides an elegant
assumption over the observables itself for when the conditional i.i.d.
assumption~\eqref{eqn:conditional iid} holds.
\begin{definition}
  \label{def:exchangeable}
  A sequence $\Obs_{1:\infty}$ is infinitely exchangeable if for any finite permutation $\sigma$, we have
  \[\P(\Obs_1,\Obs_2, \Obs_3, \dots) = \P(\Obs_{\sigma(1)},\Obs_{\sigma(2)},\Obs_{\sigma(3)}, \dots).\]
\end{definition}
\begin{theorem}[De Finetti's theorem]
  If a sequence $\Obs_{1:\infty}$ is infintely exchangeable (Assumption~\ref{def:exchangeable}), then there exists a latent parameter $\theta$ and a measure $\pi(\cdot)$ over it, such that
  \begin{equation}
    \label{eqn:definetti-rep}
    \P(\Obs_{1:\infty} = \obs_{1:\infty}) = \int \prod_{t=1}^\infty \P(\Obs_t =
    \obs_t \mid \theta) \pi(d \theta).
  \end{equation}
\end{theorem}

A key structural property that this perspective highlights is that the
one-step probabilities~\eqref{eqn:one-step} must correspond to posterior
predictions consistent with a single prior. Prior works study minimal
conditions that one-step probabilities should satisfy to guarantee a notion of
predictive coherence, ensuring they (roughly) follow Bayes' rule according to
some prior. For example, \citet{BertiPrRi04} discusses a notion called
\emph{conditionally identically distributed} (c.i.d.), which extends the
familiar concept of exchangeability
\begin{definition}
  \label{def:cid}
  $\Obs_{1:\infty}$ is c.i.d.\ if
\begin{equation}
  \label{eqn:cid}
  \P(\Obs_{t+2} = \obs \mid  \Obs_{1:t})
  = \P(\Obs_{t+1} = \obs \mid  \Obs_{1:t})
  \eqdef p_t(\obs \mid \Obs_{1:t})
  \eqdef p_t(\obs)~~~\mbox{for all}~\obs \in \R.
\end{equation}
In the case of covariates, given that $\Cov\sim P_\Cov$ independently, the above definition extends to $\Cov_{1:\infty},\Obs_{1:\infty}$ being c.i.d. if
\begin{align*}
  \P(\Obs_{t+2} = \obs \mid  \Cov_{t+2} = \cov,\Obs_{1:t}, \Cov_{1:t})
  &= \P(\Obs_{t+1} = \obs \mid \Cov_{t+1} = \cov, \Obs_{1:t}, \Cov_{1:t})\\
  &\eqdef p_t(\obs \mid \cov,\Obs_{1:t}, \Cov_{1:t})
  \eqdef p_t(\obs \mid \cov)~~~\mbox{for all}~\cov,\obs \in \R.
\end{align*}
\end{definition}

This definition enforces that the one step condiitonal
distribution being the same as two step conditional distri-
bution.Evidently, if a sequence is exchangeable, it satisfies
condition~\eqref{eqn:cid}.  As we will demonstrate, exchangeability of the
generated sequence (equivalently, permutation invariance of the autoregressive
sequence model) provides valid statistical inference and ensures robust
performance on downstream tasks that require uncertainty quantification
(Sections~\ref{section:length-gen} and~\ref{section:inference}).

\subsection{Autoregressive sequence modeling}

We introduce notation for autoregressive sequence models.  Consider a sequence
of observables
\begin{equation*}
  \Obs_{1:T}^i = \{\Obs_1^i, \cdots, \Obs_{T}^i \} ~~\mbox{for}~~ i =1, \ldots,
  n,
\end{equation*}
where $\Obs_t^i$ are documents (set of tokens)---basic units of
observables---that take on continuous or discrete values. As an example,
consider a question answering task, where we expect question-answer pairs to
be exchangeable. Naturally, we do not expect exchangeability at the token
level, so our subsequent discussion on Bayesian inference only applies over
\emph{units} of exchangeability (e.g., documents).

Define one-step predictive probabilities over documents produced by an autoregressive model $\what{p}$ as
\begin{equation}
  \label{eqn:one-step}
  \text{One-step autoregressive probability:}\quad \quad \what{p}_t(\obs) \defeq \what{p}_t(\obs \mid \Obs_{1:t})
  \defeq \what{\P}\left(\Obs_{t+1} = \obs \mid  \Obs_{1:t} \right).
\end{equation}
In the covariate setting, we use the following notations:
  \[\what{p}_t(\obs\mid\cov)\eqdef \what{p}_t(\obs\mid \cov,\Obs_{1:t},\Cov_{1:t})\eqdef \what{\P}(\Obs_{t+1}=\obs\mid \Cov_{t+1}=\cov,\Cov_{1:t},\Obs_{1:t})\]
The autoregressive probabilities can also be viewed as posterior predictives
of future data given past observations, a connection widely recognized in the
burgeoning literature on ICL~\cite{XieRaLiMa21, MullerHoArGrHu22, NguyenGr22,
  ZhangZhYaWa23}. We use the two terms interchangeably in the rest of the
paper.

Generative modeling fits an autoregressive model (e.g., decoder transformer)
to optimize the joint log likelihood/marginal likelihood of the observed
sequences by autoregressively breaking it down into the above mentioned
one-step probabilities
\begin{equation}
  \label{eqn:pretraining}
  \mbox{Pre-training:}~~~~~  \maximize_{\what{p}(\cdot)}
  \left\{
    \frac{1}{n} \sum_{i=1}^n \log \what{p} \left( \Obs_{1:T}^i\right)
    = \frac{1}{n} \sum_{i=1}^n \sum_{t=0}^{T-1}
    \log \what{p}_t \left( \Obs_{t+1}^i \right)
  \right\}.
\end{equation}
The objective~\eqref{eqn:pretraining} models a subset of the usual prediction
loss used to train LLMs, since our units of analysis $y$ models
\emph{documents}. As an example, the observable sequeces $\Obs_{1:T}$ could be
a list of images, each represented by a set of tokens, and we model each image
as a single unit of observation (instead of the pixels).  A generative model
$\what{p}$ can be used to tackle a range of different tasks by conditioning on
any sequence $\Obs_{1:s}$ (typically a prompt) at inference time (``in-context
learning'').


\section{Bayesian modeling \emph{a la} De Finetti}
\label{section:de-finetti}

We now demonstrate that under exchangeability, autoregressively generating
from a pre-trained model is equivalent to sampling a latent parameter from the
posterior distribution. This observation is not novel and dates back to De
Finetti: a long line of work in Bayesian statistics focuses on modeling
posterior predictive probabilities~\cite{DeFinetti33, BertiReRi98,
  FortiniLaRe00, FortiniPe14, HahnMaWa18, BertiDrPrRi21, BertiDrLePrRi22,
  FongHoWa23}, which we now reinterpret as autoregressive sequence models
trained using modern computational tools. This classical yet relatively
unknown insight allows us to extend the scope of ICL to tasks that require
quantification of epistemic uncertainty.

We wish to move away from modeling a fictitious latent parameter $\theta$ that
has no physical meaning, since it is never observed. Instead, we want to use
the sequence $\Obs_{1:\infty}$, which is observable in principle and has a
direct physical interpretation: the ``future'' sequence $\Obs_{s+1:\infty}$ is
simply \emph{yet} to be observed.  De Finetti's theorem shows that the two
seemingly different modeling viewpoints are in fact equivalent.  Its
characterization of exchangeable sequences~\cite{DeFinetti33, DeFinetti37}
goes beyond the representation from Equation~\eqref{eqn:definetti-rep}:
\citet{DeFinetti37, HewittSa55} show that the latent parameter $\theta$ in
Equation~\eqref{eqn:definetti-rep} is in fact entirely a function of the
infinite sequence of observables $\Obs_{1:\infty}$.

\begin{wrapfigure}{r}{.35\columnwidth}
  \centering
  \vspace{-.2cm}
  \hspace{0.2cm} 
  \includegraphics[scale = 0.2]{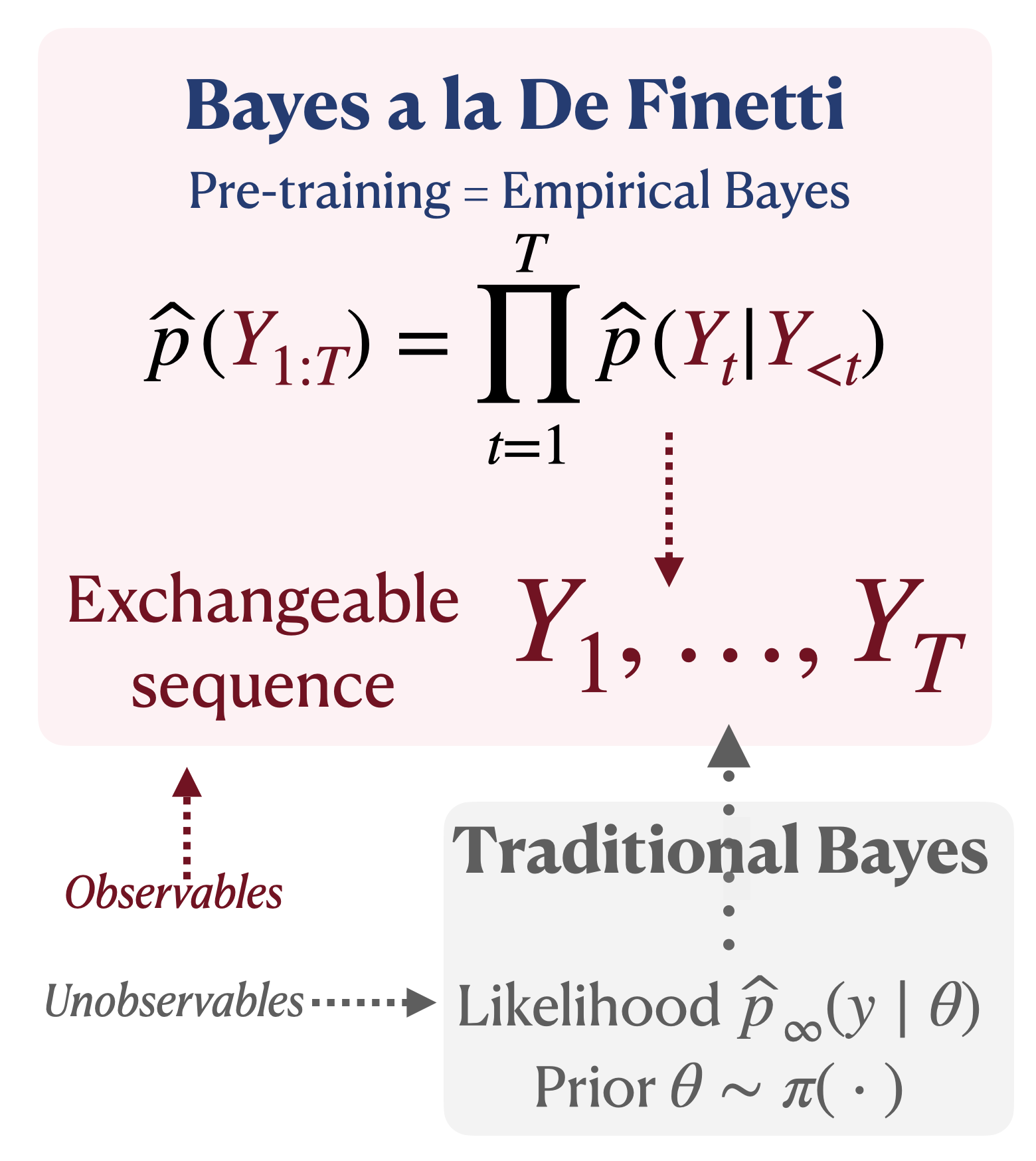}
  \vspace{-0.1cm}
  \label{fig:definetti}
\end{wrapfigure}
We are able to
conclude the following:
\begin{itbox}
  \centering
  Autoregressive probabilities~\eqref{eqn:one-step} are sufficient
  primitives for defining a  Bayesian model.
\end{itbox}
As an example~\cite{FongHoWa23}, consider the posterior predictive mean
\begin{equation*}
  \bar{\theta}_T = \E[\theta\mid\Obs_{1:T}].
\end{equation*}
By Doob's martingale convergence theorem, $\bar{\theta}_T$ converges almost
surely to a random variable
$\bar{\theta}_\infty = \theta\sim \pi(\cdot)$ that is a function of the
infinite sequence $\Obs_{1:\infty}$. Note that mathematically, the posterior
predictives are random probability measures defined over realizations of
$\Obs_{1:\infty}$. Given a random infinite dataset $\obs_{1:\infty}$, the
limiting point estimate $\bar{\theta}_\infty (\obs_{\infty})$---the
posterior mean computed on the entire dataset---is in fact distributed
according to the true prior $\pi(\cdot)$. Refer to Figure~\ref{fig:icl-inference} for a visualization.

This equivalence highlights the following fact.
\begin{itbox}
  \centering Epistemic uncertainty in $\theta$ is equivalent to predictive
  uncertainty in the long sequence $\Obs_{1:\infty}$.
\end{itbox}
De Finetti's work~\cite{CifarelliRe96} focuses on modeling the relationships
between observable quantities~\eqref{eqn:one-step}. In this view, we can
validate the modeler's claims by masking part of the observed data from the
modeler. As we explain below, this allows using validation losses on hidden
data---the empirical foundation of ML---to measure the quality of the model's
ability to comprehend uncertainty.

\subsection{Empirical Bayes and sequence modeling}

To operationalize De Finetti's philosophy, we take the \emph{posterior
  predictive probabilities}~\eqref{eqn:one-step} as our modeling primitive to
approximate the probability of seeing the observed dataset
\begin{equation*}
  \mbox{Marginal likelihood:}
\qquad 
  \P(\Obs_{1:T} = \obs_{1:T} ) \approx \what{p}(\obs_{1:T}) = \prod_{t=0}^{T-1}
\what{p}_t(\obs_{t+1})
\end{equation*}
Instead of priors and likelihoods, Bayes \emph{a la} De Finetti specifies
one-step probabilities~\eqref{eqn:one-step} on \emph{observables}. A long line
of work in Bayesian statistics advocates for this approach to Bayesian
modeling~\cite{Roberts65, Ericson69, Dawid92, DawidVo97, BertiReRi98,
  FortiniLaRe00, FortiniPe14, HahnMaWa18, BertiDrPrRi21, BertiDrLePrRi22,
  FongHoWa23}.  They propose simple parameterizations for one-step
probabilities (e.g., copulas~\cite{HahnMaWa18, FongHoWa23}) and identify
conditions under which one-step posterior predictive distributions implicitly
characterize the prior and likelihood over the latent factor $\theta$.

We note a specific connection between empirical Bayes and autoregressive
models, which allows us to perform statistical inference using sequence models
(e.g., transformers, state space models) that are multiple orders of magnitude
more expressive than model parameterizations previously considered. Since it
is difficult to specify one-step probabilities over long sequences, we model
them using modern sequence models (e.g., transformers) and adopt the empirical
Bayes philosophy: when our one-step probabilities accurately model the
data-generating distribution, masked observations will have high marginal
likelihood $\what{p}(\Obs_{s+1:T} \mid \Obs_{1:s})$. Note that this is
\emph{precisely} the original pre-training problem~\eqref{eqn:pretraining}!
We conclude that pre-training is equivalent to performing empirical Bayes by
directly optimizing posterior predictive densities.

By modeling these one-step probabilities collectively through a sequence
model, we leverage a key factor in the empirical success of language modeling:
training a differentiable loss on a flexibly parameterized model that can be
optimized using abundant data. As long as there is a wealth of previously
observed sequences ${\Obs_{1:T}^i, i= 1, \ldots, n}$, we can train any
sequence model on perplexity to learn the collection of posterior predictives.

\subsection{Explicit Bayesian inference through
  autoregressive forward sampling}
\label{section:icl}

Given a pre-trained sequence model, we can condition on any test time
observable sequence and autoregressively predicting the next observation.
Unlike prior works that interpret in-context learning as implicit Bayesian
inference by leveraging structures like hidden Markov
models~\cite{XieRaLiMa21}, we analyze ICL as \emph{explicitly} modeling the
latent parameter through autoregressive sampling. As long as we have
exchangeability, the one-step probabilities converge to a limit, which we
interpret as a ``latent parameter $\theta$'' fully determined by an infinite
sequence of observations.  \citet{BertiPrRi04} shows that if
condition~\eqref{eqn:cid} holds, then $\{p_t\}_{t=1}^\infty$ forms a
martingale:
\begin{proposition}[Martingale property]
  \label{prop:martingale}
  Under condition~\eqref{eqn:cid}, the sequence $\{p_t\}_{t=1}^\infty$ is a
  martingale.
  \begin{equation*}
    \E[p_t(\obs) \mid \Obs_{1:t-1}] = \int \P(\Obs_{t+1} = y\mid \Obs_{1:t})
    dp(\Obs_t\mid\Obs_{t-1}) = p_{t-1}(\obs)~~~\mbox{for any}~y
    \end{equation*}
    and the martingale convergence theorem yields
\begin{equation}
  \label{eqn:mct} \exists ~\mbox{random distribution}~p_{\infty}(\cdot \mid
  \Obs_{1:\infty}) ~\mbox{s.t.}~\forall y~~ p_t(\obs \mid \Obs_{1:t}) \to
  p_{\infty}(\obs \mid \Obs_{1:\infty}) ~~\mbox{almost surely}.
\end{equation} 
In the covariate setting, the martingale property holds for the sequence $\{p_t(\obs\mid \cov)\}_{t=1}^\infty$ for any $\cov,\obs$, and the limit converges to some $p_\infty(\obs\mid\cov)$. Refer to Appendix~\ref{sec:martingale-covariate} for datail.
\end{proposition}
Observe that for any given $\obs$, $p_\infty(y)$ is a random variable, as it can can be viewed as a funciton that maps each instantiation $\obs_{1:\infty}$ to a scalar. We interpret the random limit $p_{\infty}$ as a ``latent parameter'' entirely
determined by infinite observations -- each instantiation of infinite observations $\obs_{1:\infty}$ corresponds to an instantiation of the latent parameter $\theta$. Therefore, the ICL
paradigm---conditioning on a sequence of observables and autoregressively
generating the future---is equivalent to modeling the latent parameter
$\theta := p_{\infty}$.  This is the key insight of this work: \textbf{forward
  generation is equivalent to Bayesian inference on $\theta$ \emph{a la} De
  Finetti.} Refer to Figure~\ref{fig:definetti-overview} for a conceptual example of this equivalence.

Moreover, under condition~\eqref{eqn:cid}, the pre-training
objective~\eqref{eqn:pretraining} (perplexity) is the correct performance
measure capturing Bayesian inferential capabilities.

\begin{remark}
  \label{remark: notation}
  In the following sections, we use the following notations interchangeably:
  \begin{equation*}
    \what{p}_\infty(\cdot\mid \Obs_{1:\infty}) \equiv \what{p}_\infty(\cdot\mid \theta) \equiv \what{p}_\theta(\cdot)\equiv \theta.
  \end{equation*}
  \begin{equation*}
    \what{p}_\infty(\cdot\mid \cov,\Obs_{1:\infty},\Cov_{1:\infty}) \equiv \what{p}_\infty(\cdot\mid \cov,\theta) \equiv \what{p}_\theta(\cdot\mid\cov)\equiv \theta.
  \end{equation*}
  This is due to our previous argument -- an instatiation of $\Obs_{1:\infty}$ corresponds to an instantiation of $\theta$. The notation of $\what{p}_\theta$ is often used to represent the likelihood in the traditional Bayesian setting. 
\end{remark}
\begin{assumption}
  \label{assumption:cid} The true data-generating distribution satisfies
  condition~\eqref{eqn:cid} and its pre-trained counterpart satisfies the
  analogue 
  $\what{\P}(\what{\Obs}_{t+2} = \cdot \mid \obs_{1:t}) = \what{p}_t(\cdot
  \mid \obs_{1:t})$ for all $t \in \N$ and $\obs_{1:\infty}$.
\end{assumption}

Let the filtration $\mc{F}_t$ be generated by the sequence $\Obs_{1:t}$, and
define the martingale difference sequence
\begin{equation*}
  D_t := \log \what{p}_{t}(\Obs_{t+1}\mid \Obs_{1:t})- \int p_t(\obs) \log
  \what{p}_t(\obs) d\obs.
\end{equation*}
\begin{theorem}
  \label{thm:perplexity-limit}
  Let Assumption~\ref{assumption:cid} hold, $\E[D_t^2] < \infty$, and
  \begin{equation}
    \label{eqn:regularity}
    \sum_{t=1}^\infty \P\left(|D_t |>\frac{t}{\log\log t}
      ~\Big\vert~ \mathcal{F}_{t-1}\right)
    < \infty 
    ~~\mbox{and}~~\sqrt{\sum_{j=1}^t \E\left[D_j^2\mid \mathcal{F}_{j-1}\right]
      \cdot t\log\log t}\to 0 ~~~\mbox{almost surely.}
  \end{equation}
If $p_t(\obs) \log \what{p}_t(\obs)$
  is point-wise bounded by some integrable function,
\[\frac{1}{T} \sum_{t=1}^T \log \what{p}_{t-1}(\Obs_t) \to \int p_\infty(\obs) \log
  \what{p}_\infty(\obs) d\obs \eqdef \purp(\what{p}) ~~a.s..\]
\end{theorem}
\noindent Note that the limiting quantity $H(\what{p})$ in the below result is
actually a random variable. For each instantiation of
$\Obs_{1:\infty} \iff \theta$, we have a different value of
$H(\what{p}(\mid \obs_{1:\infty}))$, the almost sure statement is over the
randomness of this instantiation. 

We give the proof of Theorem~\ref{thm:perplexity-limit} in
Appendix~\ref{sec:proof-perplexity-limit}.  In contextual tasks (e.g.,
question answering) where $\Cov$ is the random variable representing the
context/covariates (question) and $\Obs$ (answer) is generated by
$p_{\infty}(\Obs | \Cov)$, we have an analogous result:
\begin{equation*}
  \frac{1}{T} \sum_{t=1}^T \log \what{p}_{t-1}(\Obs_t\mid \Cov_{1:t}, \Obs_{1:t-1}) \to\E_{\Cov \sim P_\Cov}\left[
  \int p_\infty(\obs\mid \Cov) \log
  \what{p}_\infty(\obs\mid \Cov) d\obs \right]~~a.s..
\end{equation*}
Again, the RHS $p_\infty$ and $\what{p}_\infty$ are random variables, and the
almost sure statement is over the randomness of
$\Cov_{1:\infty},\Obs_{1:\infty}$.  We provide analytical examples for a toy
attention model in Appendix~\ref{section:transformers}.  

Jensen's inequality implies that the true
distribution is clearly the ``best model'': for any $\what{p}$,
\begin{subequations}
  \label{eqn:true-is-best}
  \begin{align}
    \label{eqn:finite-T}
    \E\left[\frac{1}{T} \sum_{t=1}^T \log
    \what{p}_{t-1}(\Obs_t)\right] & \le \E\left[\frac{1}{T} \sum_{t=1}^T \log
                                    p_{t-1}(\Obs_t)\right] \\
    \label{eqn:inf-T}
    \E_{\Obs_{1:\infty}}\left[\int p_\infty(\obs\mid \Obs_{1:\infty}) \log
\what{p}_\infty(\obs\mid \Obs_{1:\infty}) d\obs \right]&\le \E_{\Obs_{1:\infty}}\left[\int p_\infty(\obs\mid \Obs_{1:\infty}) \log p_\infty(\obs\mid \Obs_{1:\infty}) d\obs\right],
  \end{align}
\end{subequations}
where the final line can be rewritten using the shorthand
$\E_{\theta}\left[\purp(\what{p}_\theta) \right]\le
\E_{\theta}\left[\purp(p_\theta)\right]$.


\section{Length generalization}
\label{section:length-gen}

We highlight overlooked applications of ICL that require uncertainty
quantification and Bayesian reasoning. First, we demonstrate that Bayesian
inference enables \emph{length generalization}~(\ref{fig:length}) in sequence predictions,
providing robustness to longer sequence prediction contexts than those
encountered during training.  Sequences in the pre-training data will be
naturally limited in their length---there is a bound on the number of
documents $Y_t$ observed from the same exchangeable cluster. We are interested
in the model's ability to achieve robust predictive performance over sequences
longer than those seen during pre-training. This ability requires the model to
correctly propagate epistemic uncertainty, meaning it must learn to extract
information from the longer context sequence and place more confidence in its
own predictions (Figure~\ref{fig:length}).  The environment generates
observable sequences of length $T$, and we use our model, denoted as
$\what{p}$, to predict the probability of observing this sequence,
i.e. $\what{p}(\obs_{1:T})$.  Recently,~\citet{HollmannMuEgHu23} empirically
observed that transformers pre-trained on synthetic data can achieve
competitive predictive performance on real data, even over sequences longer
than that seen during pre-training.

\begin{figure}
  \begin{minipage}[b]{0.49\textwidth}
    \centering
    \begin{tikzpicture}[
        >=latex,
        every node/.style={font=\footnotesize{}},
      ]

      \node[rectangle,minimum height=4ex] (model) {};

      \newcommand{\nudgexskip}{0.5em};
      \newcommand{\smolxskip}{1.2em};
      \newcommand{\lorgxskip}{3.0em};

      \coordinate (outputs) at ($(model) + (up:6.5ex)$);
      \coordinate (inputs x) at ($(model) + (down:10.5ex)$);
      \coordinate (inputs y) at ($(inputs x) + (up:3.0ex)$);

      \coordinate (input) at (inputs x);
      \coordinate (input begin) at ($(input) + (right:\smolxskip)$);

      \coordinate (input) at ($(input |- inputs x) + (right:\smolxskip)$);
      \node[anchor=base] at (input) (input) {$x_1$};
      \coordinate (x1) at (input.base);
      \draw[->] (input) -- (input |- model.south);

      \draw[->] (x1 |- model.north) -- (x1 |- outputs)
          node[anchor=south] (py1) {$\hat{p}(\hat{y}_1)$};

      \coordinate (input) at ($(input |- inputs y) + (right:\smolxskip)$);
      \node[anchor=base] at (input) (input) {$y_1$};
      \coordinate (y1) at (input.base);
      \draw[->] (input) -- (input |- model.south);

      \coordinate (input) at ($(input |- inputs x) + (right:\smolxskip)$);
      \node[anchor=base] at (input) (input) {$x_2$};
      \coordinate (x2) at (input.base);
      \draw[->] (input) -- (input |- model.south);

      \draw[->] (x2 |- model.north) -- (x2 |- outputs)
          node[anchor=south] (py2) {$\hat{p}(\hat{y}_2)$};

      \coordinate (input) at ($(input |- inputs y) + (right:\smolxskip)$);
      \node[anchor=base] at (input) (input) {$y_2$};
      \coordinate (y2) at (input.base);
      \draw[->] (input) -- (input |- model.south);

      \coordinate (input) at ($(input |- inputs x) + (right:\lorgxskip)$);
      \node[anchor=base] at (input) (input) {$x_{T_{pt}}$};
      \coordinate (xtpt) at (input.base);
      \draw[->] (input) -- (input |- model.south);

      \draw[->] (xtpt |- model.north) -- (xtpt |- outputs)
          node[anchor=south] (pytpt) {$\hat{p}(\hat{y}_{T_{pt}})$};

      \coordinate (input) at ($(input |- inputs y) + (right:\smolxskip) + (right:\nudgexskip)$);
      \node[anchor=base] at (input) (input) {$y_{T_{pt}}$};
      \coordinate (ytpt) at (input.base);
      \draw[->] (input) -- (input |- model.south);

      \coordinate (input) at ($(input |- inputs x) + (right:\lorgxskip)$);
      \node[anchor=base] at (input) (input) {$x_{T-1}$};
      \coordinate (xt1) at (input.base);
      \draw[->] (input) -- (input |- model.south);

      \draw[->] (xt1 |- model.north) -- (xt1 |- outputs)
          node[anchor=south] (pyt1) {$\hat{p}(\hat{y}_{T-1})$};

      \coordinate (input) at ($(input |- inputs y) + (right:\smolxskip) + (right:\nudgexskip)$);
      \node[anchor=base] at (input) (input) {$y_{T-1}$};
      \coordinate (yt1) at (input.base);
      \draw[->] (input) -- (input |- model.south);

      \coordinate (input) at ($(input |- inputs x) + (right:\smolxskip) + (right:\nudgexskip)$);
      \node[anchor=base] at (input) (input) {$x_{T}$};
      \coordinate (xt) at (input.base);
      \draw[->] (input) -- (input |- model.south);

      \draw[->] (xt |- model.north) -- (xt |- outputs)
          node[anchor=south] (pyt) {$\hat{p}(\hat{y}_{T})$};

      \coordinate (input end) at (input);

      \node[rectangle,rounded corners, fill=black!30,
            minimum width=18em,  
            minimum height=4ex,
            align=center,
            ]
        at ($(input begin |- model)!0.5!(input end |- model)$)
        {\textbf{\normalsize{}Pre-trained Sequence Model}};

      \path (y2.north) -- node[yshift=3ex] {\ldots} (y2.north -| xtpt.north);
      \path (ytpt.north) -- node[yshift=3ex] {\ldots} (ytpt.north -| xt1.north);

      \node[rectangle,dashed,
            draw,
            inner sep=0pt,
            fit=(py1) (pyt)] {};

      \draw[decorate,decoration={brace,amplitude=0.8em}]
          ($(ytpt |- x1) + (down:1.5ex)$) -- node[below,yshift=-1.5ex] {pre-trained length}
          ($(x1) + (down:1.5ex)$);

      \draw[decorate,decoration={brace,amplitude=0.8em}]
          ($(xt) + (down:1.5ex)$) -- node[below,yshift=-1.5ex] {longer sequence}
          ($(xt1) + (down:1.5ex)$);

      \node at ($(model) + (down:19ex)$) {};
    \end{tikzpicture}
    \caption{Length generalization.}
  \label{fig:length}
  \end{minipage}
  \hfill
  \begin{minipage}[b]{0.49\textwidth}
    \centering
    \begin{tikzpicture}[
        >=latex,
        every node/.style={font=\footnotesize{}},
      ]

      \node[rectangle,minimum height=4ex] (model) {};

      \newcommand{\smolxskip}{0.8em};
      \newcommand{\medxskip}{1.2em};
      \newcommand{\lorgxskip}{2.2em};

      \coordinate (outputs) at ($(model) + (up:6.5ex)$);
      \coordinate (inputs x) at ($(model) + (down:10.5ex)$);
      \coordinate (inputs y) at ($(inputs x) + (up:3.0ex)$);

      \coordinate (input) at (inputs x);
      \coordinate (input begin) at ($(input) + (right:\smolxskip)$);

      \coordinate (input) at ($(input |- inputs x) + (right:\smolxskip)$);
      \node[anchor=base] at (input) (input) {$x_1$};
      \coordinate (x1) at (input.base);
      \draw[->] (input.north) -- (input |- model.south);

      \coordinate (input) at ($(input |- inputs y) + (right:\smolxskip)$);
      \node[anchor=base] at (input) (input) {$y_1$};
      \coordinate (y1) at (input.base);
      \draw[->] (input.north) -- (input |- model.south);

      \coordinate (input) at ($(input |- inputs x) + (right:\smolxskip)$);
      \node[anchor=base] at (input) (input) {$x_2$};
      \coordinate (x2) at (input.base);
      \draw[->] (input) -- (input |- model.south);

      \coordinate (input) at ($(input |- inputs y) + (right:\smolxskip)$);
      \node[anchor=base] at (input) (input) {$y_2$};
      \coordinate (y2) at (input.base);
      \draw[->] (input) -- (input |- model.south);

      \coordinate (input) at ($(input |- inputs x) + (right:\lorgxskip)$);
      \node[anchor=base] at (input) (input) {$x_s$};
      \coordinate (xs) at (input.base);
      \draw[->] (input) -- (input |- model.south);

      \coordinate (input) at ($(input |- inputs y) + (right:\smolxskip)$);
      \node[anchor=base] at (input) (input) {$y_s$};
      \coordinate (ys) at (input.base);
      \draw[->] (input) -- (input |- model.south);

      \coordinate (input) at ($(input |- inputs x) + (right:\medxskip)$);
      \node[anchor=base] at (input) (input) {$x^b_{s+1}$};
      \coordinate (xbs1) at (input.base);
      \draw[->] (input) -- (input |- model.south);

      \draw[->] (xbs1 |- model.north) -- (xbs1 |- outputs)
          node[anchor=south] (pybs1) {$\hat{p}(\hat{y}^b_{s+1})$};

      \coordinate (input) at ($(input |- inputs y) + (right:\lorgxskip)$);
      \node[anchor=base] at (input) (input) {$y^b_{s+1}$};
      \coordinate (ybs1) at (input.base);
      \draw[->] (input) -- (input |- model.south);

      \draw[->,draw=blue] ($(pybs1.south) + (right:0.5em)$) |-
          node[right,pos=0.1] {\color{blue}$\sim$} (input.west);

      \coordinate (input) at ($(input |- inputs x) + (right:\medxskip)$);
      \node[anchor=base] at (input) (input) {$x^b_{s+2}$};
      \coordinate (xbs2) at (input.base);
      \draw[->] (input) -- (input |- model.south);

      \draw[->] (xbs2 |- model.north) -- (xbs2 |- outputs)
          node[anchor=south] (pybs2) {$\hat{p}(\hat{y}^b_{s+2})$};

      \coordinate (input) at ($(input |- inputs y) + (right:\lorgxskip)$);
      \node[anchor=base] at (input) (input) {$y^b_{s+2}$};
      \coordinate (ybs2) at (input.base);
      \draw[->] (input) -- (input |- model.south);

      \draw[->,draw=blue] ($(pybs2.south) + (right:0.5em)$) |-
          node[right,pos=0.1] {\color{blue}$\sim$} (input.west);

      \coordinate (input) at ($(input |- inputs x) + (right:\lorgxskip)$);
      \node[anchor=base] at (input) (input) {$x^b_T$};
      \coordinate (xbt) at (input.base);
      \draw[->] (input) -- (input |- model.south);

      \draw[->] (xbt |- model.north) -- (xbt |- outputs)
          node[anchor=south] (pybt) {$\hat{p}(\hat{y}^b_T)$};

      \coordinate (input end) at (input);

      \node[rectangle,rounded corners, fill=black!30,
            minimum width=18em,  
            minimum height=4ex,
            align=center,
            ]
        at ($(input begin |- model)!0.5!(input end |- model)$)
        {\textbf{\normalsize{}Pre-trained Sequence Model}};

      \path (y2) -- node[yshift=3ex] {\ldots} (y2 -| xs);
      \path (ybs2) -- node[yshift=3ex] {\ldots} (ybs2 -| xbt);

      \draw[decorate,decoration={brace,amplitude=0.8em}]
          ($(ys |- x1) + (down:1.5ex)$) -- node[below,yshift=-1.5ex] {context}
          ($(x1) + (down:1.5ex)$);

      \draw[decorate,decoration={brace,amplitude=0.8em}]
          ($(xbt |- x1) + (down:1.5ex)$) -- node[below,text width=10em,align=center,yshift=-1.5ex] {autoregressively \\ generate $B$ times}
          ($(xbs1) + (down:1.5ex)$);

      \node at ($(model) + (down:19ex)$) {};
    \end{tikzpicture}
    \caption{Statistical inference.}
    \label{fig:inference}
  \end{minipage}
\end{figure}

In this work, we theoretically analyze the performance gap between the ground
truth environment $Q$ and the model, expressed as
${\E}_{\Obs_{1:T}\sim Q} [\log q(\Obs_{1:T}) - \log \what{p}(\Obs_{1:T})]$,
and demonstrate that the model's performance on longer sequences is
characterized by the limiting perplexity $H(\what{p})$ defined in
Theorem~\ref{thm:perplexity-limit}.
Assuming that our sequence model $\what{p}$ is infinitely exchangeable, De
Finetti's theorem shows that there is a prior $\pi$ and a likelihood
$\what{p}_{\infty}$ corresponding to this model.  In this section, we will
require a stronger condition, that data generated from $\what{p}$ is mixture
of i.i.d.\ over a \emph{finite-dimensional} latent parameter $\theta \in \R^d$.
Concretely, we let
$\theta\mapsto \what{p}_{\infty}(y \mid \theta) =: \what{p}_{\theta}(y),
\theta\in\Theta : = \mbox{supp}(\pi)$ be the likelihood function mappings
implicitly defined by the sequence model $\what{p}$. At test time, the
environment generates i.i.d.\ sequences $\Obs_{1:T}\sim Q$, where the
likelihood under $Q$ does not necessarily lie in the likelihood class
$\{\what{p}_\theta: \theta\in\Theta\}$ that the model posits.

\subsection{ICL generalizes robustly if exchangeable}
\label{section:seq-prediction}

By using an analogous mathematical machinery used to prove the Bernstein-von
Mises theorem~\cite{VanDerVaart98, KleijnVa12}, we can characterize the
predictive performance of the model under very long sequences. In particular,
the limiting performance gap achieves the best in-class performance among the
set of possible likelihoods $\Theta$. We assume the KL divergence of the model
$\what{p}$ relative to $q$ is finite and $\theta\opt \in \Theta$ is the KL
projection of the i.i.d.\ data-generating distribution $Q(Y = \cdot)$ to the
likelihood space $\{\what{p}_{\theta}: \theta \in \Theta\}$
\begin{equation*}
  \theta\opt = \argmin_{\theta \in \Theta} \dkl{Q(Y =
  \cdot)}{\what{p}_{\theta}(Y = \cdot)}.
\end{equation*}
Define the standardized score function
\begin{equation*}
  S_T(\theta) := \frac{1}{\sqrt{T}}\sum_{t=1}^T
  \dot{\ell}_{\theta}(\Obs_t)
  ~~\mbox{where}~~
  \dot{\ell}_{\theta}(\obs)
  := \nabla_\theta \log \what{p}_\theta(\obs).
\end{equation*}

In the following result, assume the density $\what{p}_\theta(\obs)$ is twice
continuously differentiable at $\theta\opt$ for almost every $\obs$, and let
$\theta \mapsto \dkl{Q(Y = \cdot)}{\what{P}_{\theta}(Y = \cdot)}$ have a
positive definite Hessian $V_{\theta\opt}$ at $\theta = \theta\opt$.
Moreover, assume there exists a $\delta$ such that
\begin{equation*}
  \E_{\Obs\sim Q}\left[\left|\frac{\partial}{\partial \theta_j}\log
      \what{p}_\theta(\Obs)\right|^2\right]<\infty
  ~~~\mbox{and}~~~
    \E_{\Obs\sim Q}\left[\underset{\|\theta-\theta\opt\|<\delta}{\sup}
    \left|\frac{\partial^2}{\partial
        \theta_j \partial \theta_k}
      \log \what{p}_\theta(\Obs)\right|^2\right]<\infty.
\end{equation*}
\begin{theorem}
  \label{thm:regret}
  Let the prior density $\pi$ be continuous and positive in a neighborhood of
  $\theta\opt$, and let
  $\E_{Q}[\dot{\ell}_{\theta\opt}\dot{\ell}_{\theta\opt}^T]$ be
  invertible. For $\Obs_t \simiid Q$, we have
  \begin{align*}
    & \inf_{\theta\in \Theta} \dkl{Q(Y = \cdot)}{\what{p}_\theta(Y =\cdot)} \\
    & \ge
      \underset{T \to \infty}{\limsup}\biggl(\frac{1}{T} \E_{\Obs_t\simiid Q}[\log q(\Obs_{1:T})
      - \log \what{p}(\Obs_{1:T})] \\
    & \hspace{2cm}  - \frac{d}{2T}\log\frac{T}{2\pi}
      + \frac{1}{2T}\E_{\Obs_{t}\simiid Q}[S_T(\theta\opt)^TV_{\theta\opt}^{-1}S_T(\theta\opt)]
    - \frac{1}{T}\log \frac{1}{\pi(\theta\opt)}
    - \frac{1}{2T} \log\det(V_{\theta\opt})  \biggr).
  \end{align*}
\end{theorem}
\noindent See Appendix~\ref{sec:regret-proof} for the proof.

Combining with Theorem~\ref{thm:perplexity-limit}, we have shown that an
\emph{exchangeable} model's performance is optimal in the induced model class
$\Theta$ up to $O(\frac{\log T}{T})$
\begin{align*}
  \dkl{q}{\what{p}_{\infty}}
  \le 
  \inf_{\theta\in \Theta} \dkl{Q(Y = \cdot)}{\what{p}_\theta(Y =\cdot)}
  + O\left(\frac{1}{T}\left( \log T + \log \frac{1}{\pi(\theta\opt)} \right) \right).
\end{align*}
For contextual tasks where $\Cov$ is the covariates and $\Obs$ is the
observable, we have the analogous result:
\begin{align*}
  \label{eqn:combination-result}
  \underset{\Cov\sim P_\Cov}{\E}&\left[\dkl{q(\cdot\mid\Cov)}{\what{p}_{\infty}(\cdot\mid\Cov)}\right]\\
  &\le 
  \inf_{\theta\in \Theta} \underset{\Cov\sim P_\Cov}{\E}\left[\dkl{Q(\Obs = \cdot\mid \Cov)}{\what{p}_\theta(\Obs =\cdot\mid \Cov)}\right]
  + O\left(\frac{1}{T}\left( \log T + \log \frac{1}{\pi(\theta\opt)} \right) \right).
\end{align*}

This demonstrates that optimizing perplexity is the appropriate objective
function to ensure the quality of Bayesian inference for the exchangeable
model.  Our result partially explains the striking robustness of ICL against
distribution shift from the pre-training distribution.  So long as the
implicit and misspecified prior $\pi$ puts some weight on the best in-class
approximation $\theta\opt$ to the data-generating ICL environment
$Q(Y = \cdot)$, the sequence model becomes a robust predictor, incurring only
$\log T$ regret as it sees more contexts. These bounds formalize and highlight
two key components of pre-training that practitioners track: the diversity of
pre-training data (i.e., assigning weights to $\theta\opt$) and perplexity.

\subsection{Case study: a one-layer transformer}
\label{section:example-length-gen}

To solidify our insights from Theorem~\ref{thm:regret}, we provide an explicit
example where the sequence model is based on a simple one-layer transformer.
In the contextual case with $X_t \in \R^d$, we consider a model
$\what{p}_t(\Obs_{t+1}\mid \Cov_{t+1})$ that is parameterized by an
self-attention layer given by $Q, K, V \in \mathbb{R}^{d \times T}$ matrices,
followed by an extra layer on top that takes the output of the $Q,K,V$ matrices
to a $\R^{2T}$ dimensional vector, representing the mean and variances of the
posterior predictive distributions. In particular, we take the model class
$\what{p}\in \mathcal{P}$ to be the family of normal distributions.

We organize the input matrix $\mathbf{Z}_{t+1}$ which is fed into a
self-attention layer parameterized by $Q, K, V$ to give the output
$\what{\mathbf{Z}}_{t+1}\in \mathbb{R}^{d\times t+1}$
\begin{equation}
  \label{eqn:io}
\mathbf{Z}_{t+1} := \begin{bmatrix}
\Cov_1 & \dots & \Cov_{t} & \Cov_{t+1} \\
\Obs_1 & \dots & \Obs_{t} & 0,
\end{bmatrix}
\rightarrow
~\mbox{self-attention } (Q, K, V)
\rightarrow
\what{\mathbf{Z}}_{t+1}:= \begin{bmatrix}
\what{\Cov}_1 & \dots & \what{\Cov}_{t} & \what{\Cov}_{t+1} \\
\what{\Obs}_1 & \dots & \what{\Obs}_{t} & \what{\Obs}_{t+1}
\end{bmatrix}
\end{equation}
where $\what{\Cov}_t$ is the prediction at time $t$ and $\what{\Obs}_t$ is the
target at time $t$.  This output matrix then goes through the final linear
layer to get the final prediction vector
\[
  [\what{\mu}_1,\what{\sigma}_1, \dots, \what{\mu}_{t+1},\what{\sigma}_{t+1}]
  \implies \what{p}_{t}(\obs\mid\cov) =
  \mathcal{N}(\what{\mu}_{t+1}(\cov),\what{\sigma}_{t+1}^2(\cov))(\obs) \quad \forall t =
  1,\dots,T-1.
\]
Throughout, we let $\vec{\Cov}_t\in\R^{t\times d}$ be the set of contexts
stacked together (design matrix) and let $\vec{\Obs}_t\in\R^{t}$ be the vector
of target outcomes
\begin{equation*}
\vec{\Cov}_t := \begin{bmatrix}
\Cov_1 & \cdots & \Cov_{t} 
\end{bmatrix}^\top
,~~~
\vec{\Obs}_t :=  \begin{bmatrix}
\Obs_1 & \cdots & \Obs_{t} 
\end{bmatrix}^\top.
\end{equation*}

Following previous works~\cite{VonOswaldEtAl23, WuZoChBrGuBa23}, we assume for
simplicity that the final layer and the self-attention parameters $Q,K,V$
matrices are such that the output vector always has the variances matching the
closed form oracle. In other words, we only study the expressiveness of a
one-layer transformer on \emph{mean} estimation.  Elementary
calculations---which we give in Appendix~\ref{proof:gd}---show that
autoregressive mean predictions are equivalent to performing one-step gradient
descent on the squared loss, based on the observed data so far.
\begin{lemma}[Reparameterization]
  \label{lemma:gd}
  The autoregressive mean predictions of the transformer described above is
  given by one-step gradient descent for linear regression
  \begin{equation}
    \label{eqn:reparam}
    \what{\mu}_{t+1}(\cov) =
    \frac{1}{t+1} \vec{\Obs}_{t}^\top \vec{\Cov}_{t} \Gamma^\top
    \cov
  \end{equation}
  where $\Gamma \in \R^{d \times d}$ is a reparameterization of the trainable
  part of the transformer parameters.
\end{lemma}



\paragraph{Data-generating process} To substantiate our abstract results in
the previous subsection, we consider a tractable data-generation
distribution. We take the usual \textsc{Bayesian linear regression} problem
with latent parameters $w \sim \pi$
\begin{equation}
  \label{eqn:blr}
  w \sim N(0, \tau^2 I), ~~~\Cov_t \simiid N(0, H),
  ~~~ \Obs_t = w^\top \Cov_t + \varepsilon_t
  ~~\mbox{where}~~\varepsilon_t \simiid N(0, \sigma^2)
  ~~\mbox{for a known}~~\sigma^2.
\end{equation}
We assume $H$ is full rank throughout.
The pretraining data is generated from the distribution~\eqref{eqn:blr},
where we assume the pre-training sequence has length $T_{\rm pt}$.

At inference time, the modeler observes data generated i.i.d.~from a
particular distribution $Q$. Denoting by $w_q$ the coefficients that give test
data, we abuse notation to write $\Obs_t = w_q^\top \Cov_t + \varepsilon_t$.
In this case, the oracle autoregressive variances (which we assume the modeler
knows) are given by
\begin{equation}
  \label{eqn:posterior-var}
  \what{\sigma}^2_t = \cov_t^T
  A_{t-1}^{-1}\cov +\sigma^2
  \quad\mbox{where}\quad
  A_{t-1} \defeq \frac{1}{\sigma^2}\vec{\Cov}_{t-1}^T\vec{\Cov}_{t-1}+\frac{1}{\tau^2}I.
\end{equation}

\paragraph{Exact characterization of length generalization}

Any fixed weight $\Gamma$ for the one-layer transformer defines an
infinitely exchangeable model (since self-attention is permutation
invariant). By De Finetti's theorem, an infinitely exchangeable sequence model
implicitly defines a likelihood and prior. In the case where the oracle
posterior variance~\eqref{eqn:posterior-var} is known, we can go beyond the
approximation in Theorem~\ref{thm:regret} and exactly characterize the excess
risk of our sequence model.  The exact calculations are given in
Appendix~\ref{proof:example-blr-main}.
\begin{proposition}
  \label{prop:blr-main-term-convergence}
  Consider the test-time data generating distribution $Q$ described above and
  let $\Cov \sim P_\Cov$ be independent of everything else.
  Then, we have
  \begin{align*}
    & \frac{1}{T}
    \E_Q\left[\log q(\Obs_{1:T}\mid X_{1:T})
      -\log\what{p}(\Obs_{1:T}\mid X_{1:T})\right] \\
    & = \frac{1}{2T} \sum_{t=1}^T \E_Q \left[
      \frac{-\Cov^\top A_{t-1} \Cov}{\sigma^2 + \Cov^\top A_{t-1} \Cov}
      +  \log\left( 1 + \frac{\Cov^\top A_{t-1} \Cov}{\sigma^2}\right)\right.\\
      &\qquad \qquad \qquad\qquad\qquad\left.+ \frac{1}{\sigma^2 + \Cov^\top A_{t-1} \Cov}
      \left(w_q^\top \Cov - \frac{1}{t} \vec{\Obs}_{t-1}^\top \vec{\Cov}_{t-1} \Gamma^\top \Cov
      \right)^2
      \right] \\
    & \rightarrow
      \frac{1}{2 \sigma^2}
      w_q^\top (I - H\Gamma^\top)H (I - H \Gamma^\top)^\top w_q
      ~~~\mbox{as}~~T \to \infty.
  \end{align*}
\end{proposition}
\noindent In particular, if $\Gamma = H^{-1}$ the one-layer transformer
perfectly generalizes to long sequences.  

\paragraph{Pre-training a one-layer transformer}

Finally, we explicitly characterize the solution to the \emph{population}
version of the pre-training problem~\eqref{eqn:pretraining} as the number of
observations $n$ generated from the distribution~\eqref{eqn:blr} goes to
infinity. Assuming the variance is known as above, the (marginal) log
likelihood is equivalent to the averaged mean squared error across all
timesteps. Denoting the pre-training sequences length as $T_{pt}$, the
population level pre-training objective is given by
\begin{align}
  R_{T_{\rm pt}}(\Gamma)
   = \frac{1}{T_{\rm pt}} \sum_{t=0}^{T_{\rm pt}-1}
    \E \left( \frac{1}{t+1} \vec{\Obs}_t^\top \vec{\Cov}_t \Gamma^\top  \cov_{t+1}
    - \Obs_{t+1}\right)^2.
  \label{eqn:pretrain-risk}
\end{align}

Elementary calculations (e.g.,~\citet[Appendix C]{WuZoChBrGuBa23}) show  
the population-level pretraining objective~\eqref{eqn:pretrain-risk} can be
simplified 
\begin{equation*}
  R_{T_{\rm pt}}(\Gamma) = \frac{1}{T_{\rm pt}} \sum_{t=0}^{T_{\rm pt}-1}
  H^\top (\Gamma - \widetilde{\Gamma}_t)
  \widetilde{H}_t (\Gamma -\widetilde{\Gamma}_t)^\top
  + H^\top \left( \tau^2I
    -\frac{1}{T_{\rm pt}} \sum_{t=0}^{T_{\rm pt}-1}
    \widetilde{\Gamma}_t \widetilde{H}_t
    \widetilde{\Gamma}_t^\top \right)
  + \sigma^2.
\end{equation*}
where recalling $H \defeq \E[\Cov_t \Cov_t^\top]$,  we used the shorthand
\begin{align*}
  \widetilde{H}_t
  & \defeq \E\left(\frac{1}{t}\vec{\Cov}_t^\top
    \vec{\Obs}_t\right)
    \left(\frac{1}{t}\vec{\Cov}_t^\top
    \vec{\Obs}_t\right)^\top
    = \tau^2 H\left( \frac{\tr(H)+\sigma^2/\tau^2}{t} I
    +\frac{t+1}{t}H\right), \\
  \widetilde{\Gamma}_t
  & \defeq t \left( \tr(H)+\frac{\sigma^2}{\tau^2} I
    + (t+1) H\right)^{-1}
    ~~\mbox{is the least quares solution at timestep}~t.
\end{align*}
The pre-training problem can be interpreted as a multi-task learning problem
where we learn a single $\Gamma$ for all timesteps. Solving the quadratic
objective explicitly, we arrive at
\begin{equation}
  \label{eqn:optimal-gamma}
  \Gamma_{T_{\rm pt}}\opt \defeq
  \left(\sum_{t=0}^{T_{\rm pt}-1} \widetilde{H}_t \right)^{-1}
    \cdot \sum_{t=0}^{T_{\rm pt}-1} \widetilde{H}_t \widetilde{\Gamma}_t.
\end{equation}
Plugging $\Gamma_{T_{\rm pt}}\opt$ into Proposition~\ref{prop:blr-main-term-convergence}, we have
characterized the approximation error incurred by the one-layer transformer
model in the Bayesian linear regression setting.




\section{In-context learning as a Bayesian statistician}
\label{section:inference}

\begin{wrapfigure}{r}{.4\columnwidth}
  \vspace{-0.3cm} 
  \centering
  \begin{minipage}{0.35\textwidth}
    \begin{algorithmic}
      \State \textbf{Given} $(\obs_{1:s}), \what{p}_{1:\infty}, T \gg s$
      \vspace{0.1cm}
      \For{$b \rightarrow 1 \textbf{ to } B$}
          \vspace{0.1cm}
          \For{$t \rightarrow s+1 \textbf{ to } T$}
              \vspace{0.1cm}
              \State $\what{\Obs}_t^b \stackrel{\text{iid}}{\sim} \what{p}_{t-1}(\cdot\mid \obs_{1:t-1})$
              \vspace{0.1cm}
          \EndFor
          \vspace{0.1cm}
          \State $ \tau_T^b := \int g(\obs) dF^b_T(\obs)$
          \vspace{0.1cm}
      \EndFor
      \vspace{0.1cm}
      \State \textbf{Output}  $\tau_T^1, \dots, \tau_T^B \stackrel{\text{iid}}{\sim} \Pi_T(.|y_{1:s})$
    \end{algorithmic}
    \captionof{algorithm}{Autoregressive bootstraps. $F^b_T$ is the empirical distribution of $(\obs_{1:s},\what{\Obs}_{s+1:T}^b)$.}
    \label{alg:bayesian-bootstrap-inference}
  \end{minipage}
  \vspace{-0.2cm}
\end{wrapfigure}
We now explicitly highlight that ICL can be utilized for statistical
inference~(\ref{fig:inference}), extending its utility beyond typical predictive applications. This is an instantiation of our key insight: forward generation
is equivalent to Bayesian inference on $\theta$ a la De Finetti. We
study how we can use forward generation to construct confidence intervals for
a latent parameter that governs data generation.  Going back to the case
without contexts for simplicity, for some measurable function $g: \R \to \R$,
let the estimand be its mean under $p_\infty$
\begin{equation*}
  \tau\opt = \int g(\obs) p_\infty(\obs) d\obs.
\end{equation*}
Note 
Formally, $\tau$ is viewed as a random variable that depends on the infinite observable sequence $\obs_{1:\infty}$, which, equivalently, represents the environment's realization of the latent variable $\theta$. For a finite predictive observable size $T$, we can write the parameter
estimate as
\begin{equation*}
  \tau(\obs_{1:T}) = \frac{1}{T}\sum_{t=1}^T g(\obs_t) = \int g(\obs)
  dF_T(\obs)
\end{equation*}
where $F_T$ denotes the empirical distribution. Note that this approach
straightforwardly generalizes to loss minimization settings
$\argmin_\tau \int \ell(\tau,\obs) dp_\infty(y)$; for instance, in the
Bayesian linear regression setting, the parameter of interest is often the
linear coefficient.

Letting $\obs_{1:s}$ be the observed data, our goal is to generate a
confidence/credible interval around $\tau\opt$. To this end, we can draw
insights from the extensive body of work in Bayesian statistics. For instance,
\citet{EfronTi93} proposed the Bayesian bootstrap, and \citet{FongHoWa23}
introduced a clear forward sampling algorithm for the Bayesian bootstrap
within De Finetti's framework. We instantiate their approach using
autoregressive sequence models in
Algorithm~\ref{alg:bayesian-bootstrap-inference} and Figure~\ref{fig:length}.
The following classical result shows autoregressive forward sampling allows
sampling from the true posterior.

\begin{theorem}[{\citet[Lemma 2.1, Theorem 2.2]{BertiPrRi04}}]
  \label{thm:coverage}
  Let Assumption~\ref{assumption:cid} hold, and let $\what{\Obs}_{s+1:T}$ be
  autoregressively generated conditional on $\Obs_{1:s} = \obs_{1:s}$. When
  $\E[|g(\obs)|]<\infty$, we have
  \[ \frac{1}{T-s}\sum_{t=s+1}^{T} \mathbbm{1}(\what{\Obs}_t \leq \obs) \cas
    \what{P}_\infty(\obs\mid \Obs_{1:s} = \obs_{1:s}) \]
    \[\frac{1}{T-s}\sum_{t=s+1}^{T} g(\what{\Obs}_t) \cas
    \int g(\obs) \what{p}_\infty(\obs\mid \Obs_{1:s} = \obs_{1:s}) d\obs.
    \]
    Here, the almost sure (a.s.) statement applies to the remaining realization $\Obs_{s+1:\infty}$, capturing the randomness in the autoregressive generation of $\what{\Obs}_{s+1:T}$, $T\to\infty$.
  \end{theorem}

We can again show the limiting sequence loss $H(\what{p})$ offers pathwise
control over $\what{\tau}_T - \tau\opt$, where
$\what{\tau}_T = \int g(\obs) d\what{F}_T(\obs)$ is the parameter estimate
from the autoregressive generation from the model $\what{p}$.
\begin{align*}
     \lim_{T \to \infty} \what{\tau}_T - \tau\opt  \lesssim
  \norm{g}_{\infty} \sqrt{\dkl{p_\infty}{\what{p}_\infty}} \propto
  \purp(\what{p})^\half ~~\mbox{a.s.}
\end{align*}
Thus, the pre-training problem~\eqref{eqn:pretraining} is the ``right''
objective to guarantee the quality of Bayesian inference.  Under
exchangeability, autoregressive generation gives a natural Bayesian inference
procedure based on the bootstrap: calculating a $\alpha$-confidence interval
from Algorithm~\ref{alg:bayesian-bootstrap-inference} by finding the
corresponding quantile $\what{q}_{\alpha,n,B}$, we have
$p_\infty(\tau\opt \le \what{q}_{\alpha,s,B}) \sim \alpha$.

If we only care about the squared loss (as opposed to log likelihoods), the
limiting sequence loss $H(\what{p})$ in Theorem~\ref{thm:perplexity-limit}
also governs the sequential \emph{prediction} performance over long
horizons. That is, evaluating on squared loss instead of the perplexity.
Given a ``prompt'' consisting of the sequence of tokens $\Obs_{1:s}$ at
inference time, we generate $\what{\Obs}_{s+1:T}$ to predict unseen
observations $\Obs_{s+1:T}$. We evaluate ourselves on the $T$-horizon squared
loss
$\what{R}_T \defeq \frac{1}{T} \sum_{t = 1}^T (\what{\Obs}_t - \Obs_t)^2$. The
limiting perplexity $H(\what{p})$ again controls length generalization
capabilities of the fitted model
\begin{align}
  \label{eqn:regret-contextual}
  \lim_{T \to \infty} \what{R}_T  \le \var \left(
  \what{\Obs}_\infty\right) + \var \left(\Obs_\infty\right) +
\dkl{p_\infty}{\what{p}_\infty} \propto 
  (1+\purp(\what{p}))~~\mbox{a.s.}.
\end{align}


\section{Inductive biases for exchangeability}
\label{section:experiments}

Our theory highlights exchangeability/permutation invariance as a key
property of the autoregressive model that enables Bayesian inference. In this
section, we investigate various inductive biases that can be applied to
promote permutation invariance in a transformer, and study their effectiveness
in enhancing performance on the two aforementioned tasks in
Sections~\ref{section:length-gen} and~\ref{section:inference} where understanding epistemic uncertainty is
crucial.

Following prior works on ICL~\cite{GargTsLiVa22}, we study sequence generating
processes where the oracle Bayes model that knows the true prior can be
explicitly computed. This allows us to directly compare the autoregressive
loss of the fitted sequence model to that of the optimal Bayesian model.  In
particular, we return to the \textsc{Bayesian linear regression}
problem~\eqref{eqn:blr}. Clearly, the sequence prediction loss is maximized by
the Bayesian linear regression oracle that knows the true prior
$w \sim N(0, \tau^2 I)$. By the Bayes rule, this data-generating process is
equivalent to marginalizing over latent $w\sim \pi(\cdot)$, and simply
generating data from the oracle posterior predictives
$\Obs_t \sim p(\cdot \mid \Obs_{1:t-1},\Cov_{1:t})$ iteratively over
i.i.d.\ covariates $\Cov_t \simiid P_\Cov$.

\subsection{Promoting Permutation Invariance} 

Without any constraints on the autoregressive probabilities, we cannot make
reliable predictions beyond the pre-training context length. In other words,
we lack guarantees on the quality of $\what{p}_t$ for any $t > T$. This
underscores the necessity of restricting the model class to a collection of
models that enforce some coherency condition. We investigate various inductive
biases that can be applied to promote permutation invariance in a transformer
model. Specifically, we evaluate data augmentation, loss-based regularization,
and causal masking strategies as compared various positional embeddings on
GPT2 in order to improve the coherency in the autoregressive probabilities.

\paragraph{Data augmentation}
A straightforward way to promote permutation invariance is \textsc{data
  augmentation} where we permute the order of the input data during training,
so that the autoregressive model learns to provide identical outputs
regardless of the order of the input sequence. We perform permutation-based data
augmentation to train GPT2 with and without positional embeddings.

\paragraph{Regularization}
As an alternative, we propose a novel \textsc{cid regularization} method that
forces the model to respect the exchangeability structure of the data. In
addition to the usual autoregressive loss term in the training objective, we
add a KL-regularizer that encourages the model to predict the next token in
the same way as predicting the token after the next token, inspired by
condition~\eqref{eqn:cid}.  Denoting the one-step autoregressive probability by
$\what{p}_t^i \defeq \what{p}(\Obs_{t+1}^i \mid \Cov_{1:t}^i, \Obs_{1:t}^i,
\Cov_{t+1}^i)$, we add the following
term to the usual training objective
\begin{align}
  \label{eqn:kl-regularized}
  \sum_{t=0}^{T-1}  \underbrace{\log \what{p}_t^i}_{\text{autoregressive loss}}
  + ~~ \lambda~ \cdot ~ \sum_{t=0}^{T-2}
    D_{\rm kl}\Bigg(
    \underbrace{\what{p}_t^i}_{\substack{\text{one-step}\\\text{prediction}}}
  |\!| \underbrace{
    \what{p}(\Obs_{t+2}^i = \obs_{t+1}^i \mid \Cov_{1:t}^i = \cov_{1:t}^i, \Obs_{1:t}^i = \obs_{1:t}^i, \Cov_{t+2}^i = \cov_{t+1}^i)
    }_{\begin{array}{c}
    \text{two-step prediction}  \\
         \text{(if } \obs_{t+1}^i, \cov_{t+1}^i
         \text{  were observed in two steps)}
       \end{array}}
  \Bigg),
\end{align}
where the two-step predictor $\what{p}_{t+1}$ marginalizes over next-step
predictions. We estimate the regularizer using Monte Carlo samples as
detailed in Appendix~\ref{section:regularization}.

\begin{figure}[t]
  \begin{subfigure}[b]{0.5\textwidth}
    \hspace{0.84em}
    \includegraphics[scale=0.23]{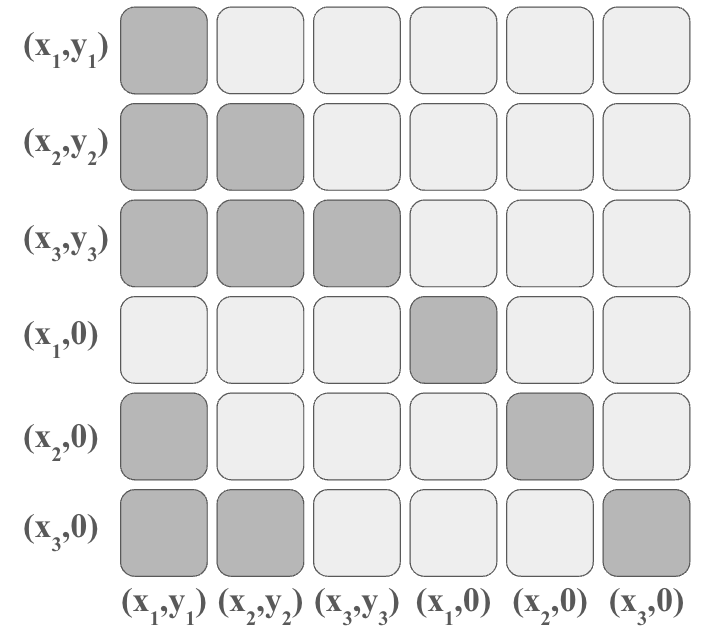}
    \caption{Attention mask}
  \end{subfigure}
  \hspace{-0.5cm}
  \begin{subfigure}[b]{0.5\textwidth}
    \centering
    \vspace{-0.5cm}
    \includegraphics[scale=0.22]{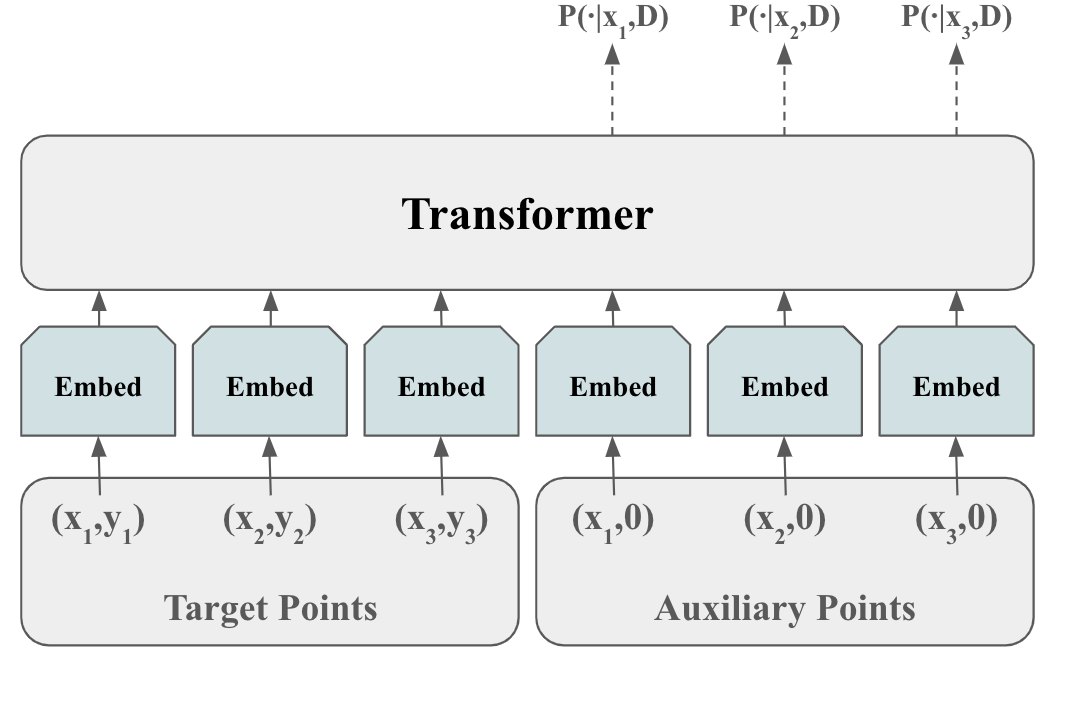}
    \caption{Architecture}
  \end{subfigure}
  \caption{The attention mask and architecture of an exchangeable transformer. Each token consists of a feature-label pair. Auxiliary tokens are target covariates padded with zeros. Both auxiliary and target tokens attend to themselves and target tokens of smaller index. Our model is then evaluated on autoregressive loss based on the predictions made on the auxiliary tokens.}
  \vspace{-.2cm}
  \label{fig:et}
\end{figure}

\paragraph{Causal masking}
Finally, we propose the \textsc{Exchangeable Transformer}, an autoregressive
transformer model without positional embeddings, an attention masks where
target tokens attend to past context, with $(x,y)$ concatenated into each
token (Figure~\ref{fig:et}). For points being predicted, the $x$ are
concatenated with zeroes, i.e., $(x,0)$. This architecture is inspired by prior
works~\cite{MullerHoArGrHu22, NguyenGr22}, but differ in a small yet important
way in how future predictions attend to prior true labels and the point being
predicted (see Appendix~\ref{section:experiments-details} for a detailed
comparison). Each token in our exchangeable transformer consists of
concatenated feature label pairs to allow for the removal of positional
embeddings. An attention mechanism and model architecture was also designed to
respect the autoregressive conditioning. See Appendix~\ref{section:experiments-details} for ablations on our architecture.

\begin{figure}[!t]
  \vspace{2cm}
  \centering    
  \begin{subfigure}[b]{0.40\textwidth}
    \centering
    \includegraphics[scale=.4]{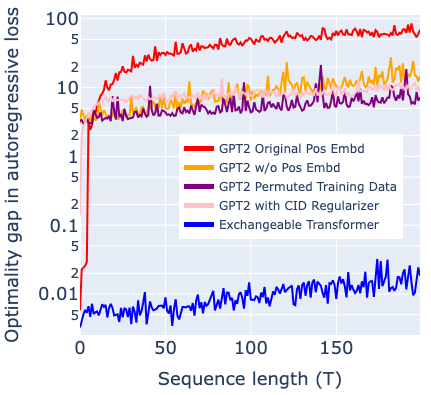}
    \caption{Bayesian linear regression}
  \end{subfigure}
  \hspace{0cm}
  \begin{subfigure}[b]{0.55\textwidth}
  \centering
    \includegraphics[scale=.17]{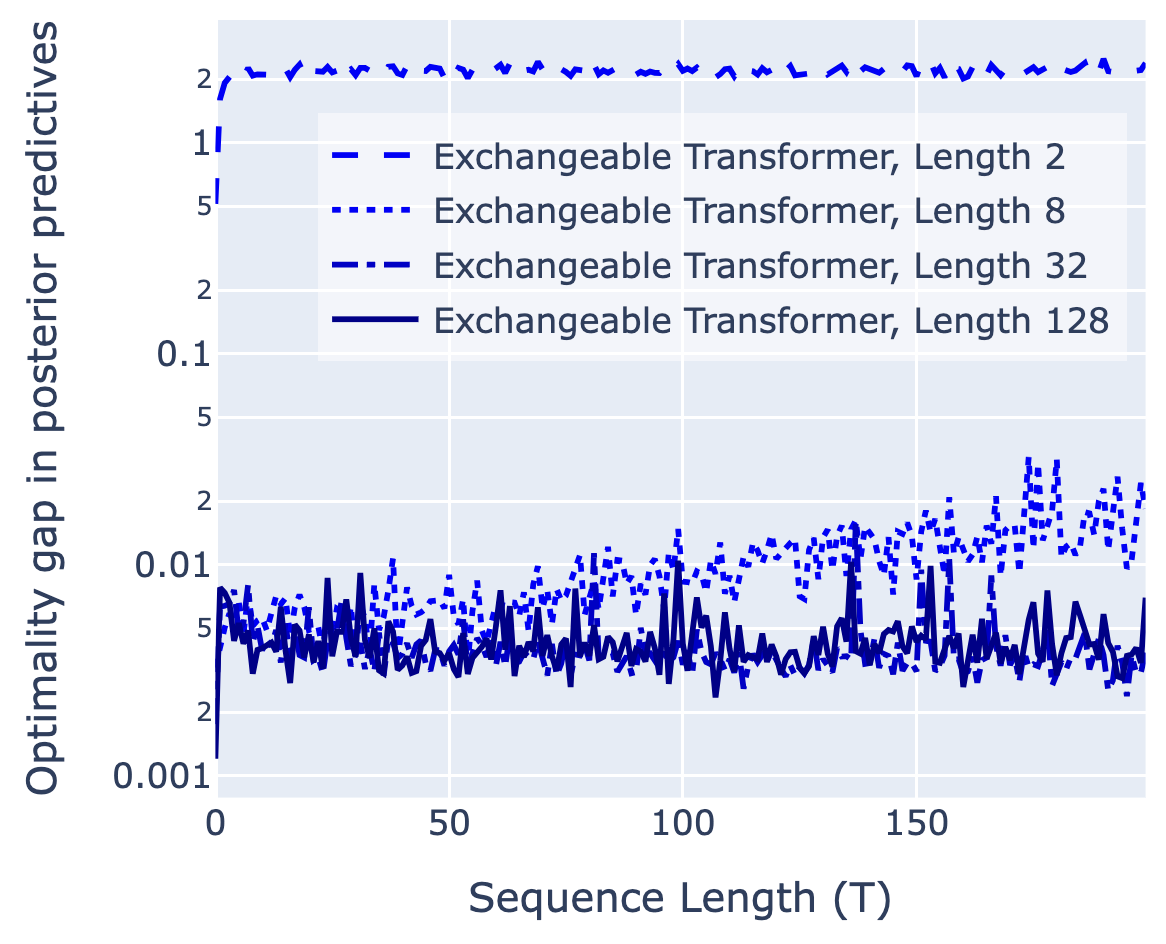}
    \caption{Exchangeable Transformer trained on different lengths}
  \end{subfigure}
  \caption{Optimality gap in autoregressive loss over sequences longer than
    that seen during pre-training,
    $\dkl{p(\cdot \mid \Cov_{1:T+1},\Obs_{1:T})}{\hat{p}(\cdot\mid
      \Cov_{1:T+1},\Obs_{1:T})}$, approximated over 100 trajectories. Both plots are in log scale.
      \textit{Exchangeable Transformer significantly
      outperforms all other approaches}.
      Compared to 9M parameters in GPT2,
      Exchangeable Transformer only has 0.2M parameters, demonstrating the
      importance of exchangeability in length generalization.
      Training on longer lengths improves posterior predictive accuracy during length generalization to a certain point. }
  \label{fig:kl}
\end{figure}

\subsection{Performance Evaluation}

\paragraph{Length generalization}

Recalling Section~\ref{section:length-gen},
we evaluate the model's ability to generalize to longer sequences than those
seen during pre-training ($T > T_{\rm pt}$). In Figure~\ref{fig:kl}, we plot
the optimality gap in autoregressive loss, which is equal to the KL
divergence between the posterior predictive under the oracle model that knows
the true prior (``DGP'') versus the fitted autoregressive sequence models. Even
when the pre-training sequence length is extremely short ($T_{\rm pt} = 8)$,
the Exchangeable Transformer generalizes well at inference time. On the other
hand, the original GPT-2 model performs poorly on longer sequences: removing
positional embeddings improves performance, meaning incorrect positional
embeddings can hinder the model's learning process, more so than having no inductive bias at all.

\paragraph{Statistical inference}

\begin{figure}[!t]
  \vspace{0.01cm}
    \centering
        \includegraphics[scale = 0.44]{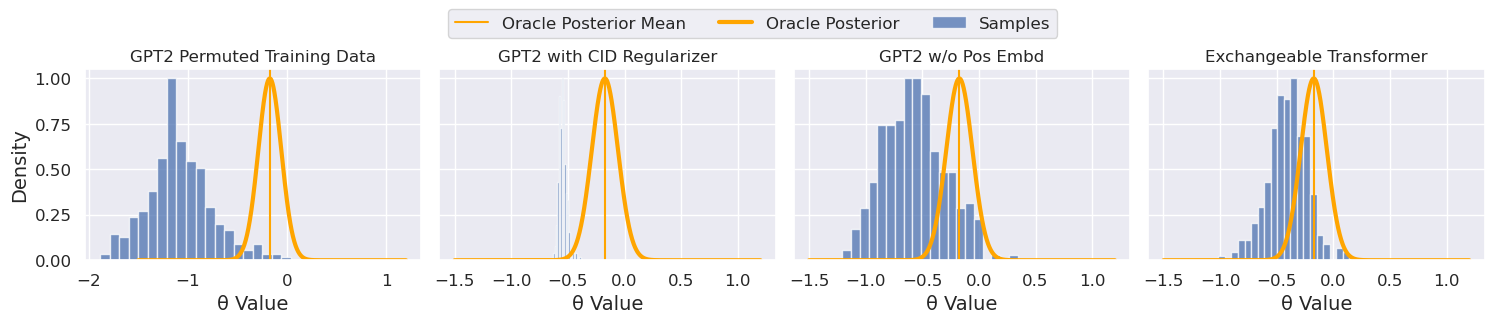}
        \caption{Approximate posterior draws from autoregressive bootstrap
          (Algorithm~\ref{alg:bayesian-bootstrap-inference}). Orange is
          oracle, Blue is the model's forward samples across 100 trajectories
          and $T=200$ forward sampling steps. The mean of the posterior is
          yellow, and the mean of the forward sample trajectories is red.}
        \label{fig:coverage}
  \hspace{1cm}
  \vspace{-0.5cm}
\end{figure}

\begin{table}[!t]
  \vspace{0.5cm}
  \centering
  \begin{tabular}{lccccc}
    \hline
     & \textbf{Dim 1} & \textbf{Dim 2} & \textbf{Dim 4} & \textbf{Dim 8} \\
    \hline
    \textbf{Length 2} & 10.653 & 7.959 & 7.188 & 12.606 \\
    \textbf{Length 8} & 0.083 & 1.087 & 3.641 & 6.187 \\
    \textbf{Length 32} & 0.025 & 0.064 & 0.167 & 0.499 \\
    \textbf{Length 128} & 0.023 & 0.048 & 0.080 & 0.187 \\
    \hline
  \end{tabular}
  \caption{Posterior Estimation Optimality Gap (KL Divergence) for Bayesian Linear Regression with Exchangeable Transformer. Training on different length and dimensions.}
  \label{tab:mytable}
\end{table}

For Bayesian linear regression where we consider statistical inference on the
latent parameter $\theta$.  Connecting to our notation in
Section~\ref{section:inference}, we have
$\theta = \tau^\star = \text{argmin}_{\theta'} \E [(\theta'^T\Cov-\Obs)^2]$
and Algorithm~\ref{alg:bayesian-bootstrap-inference} computes the ordinary
least squares estimator $\what{\theta}_b$ over autoregressively generated
trajectories. In Figure \ref{fig:coverage}, we compare the (approximate)
posterior draws of $\what{\theta}$ based on our transformers with the oracle
posterior. Recall a more precise visualization in Figure~\ref{fig:diff} where we compare the KL divergence between the oracle posterior
and the autoregressively sampled approximation. We observe a similar trend as
in Figure~\ref{fig:kl} where we see that the Exchangeable Transformer outperforms
other inductive biases by orders of magnitude, and that simply removing
positional embeddings provides a strong baseline for statistical
inference. Motivated by the strong performance of the Exchangeable
Transformer, we perform ablation studies over different covariate dimensions
and pre-training sequence lengths in Table~\ref{tab:mytable}. Our results
highlight that pre-training recipes for higher dimensional covariates is an
important direction for future work.


\section{Discussion}
\label{section:limitations}

Our work focuses on what type of uncertainty quantification sequence models
\emph{automatically} provide as a result of pre-training. Our main insight is
that autoregressive generative modeling captures epistemic uncertainty over
latent parameters that generate \emph{exchangeable} sets of documents.
Despite the classical nature of this simple insight due to De Finetti, it
appears to be not widely known in the burgeoning literature on ICL. We hope
the explicit connections we make spur subsequent methodological innovations
that expand the scope of uncertainty quantification possible by LLMs.

Our experiments are confined to synthetic settings. Scaling up inductive
biases on exchangeability will likely require substantive engineering
innovations. Our heavy reliance on De Finetti's theorem for infinitely
exchangeable sequences allows us to automatically decompose the model into
priors and likelihoods. In practice, our model can only achieve finite
exchangeability, and we have not accounted for the approximation error this
introduces in our theoretical justifications.  Addressing these errors and
extending our approach to real-world datasets are important directions for
future research.

\paragraph{Acknowledgements} We thank Daniel Russo for insightful
discussions. This work was partially supported by the CBS Digital Future
Initiative.



\newpage




\bibliographystyle{abbrvnat}

\setlength{\bibsep}{.7em}

\bibliography{../bib.bib}

\newpage
\appendix
\section{Martingale Property in Covariate Setting}
\label{sec:martingale-covariate}
Given a sequence of covariate and observable pairs $\{(\Cov_t, \Obs_t)\}_{t=1}^T$, we define the one-step predictive probability as 
\begin{equation*}
    \P(\Obs_{t+1} = \obs \mid \Cov_{t+1} = \cov, \Obs_{1:t}, \Cov_{1:t}) \eqdef p_t(\obs \mid \cov) \eqdef p_t(\obs \mid \cov,\Cov_{1:t}, \Obs_{1:t}).
\end{equation*}
We then have the martingale property
\begin{align*}
    \E[p_t(\obs\mid\cov)\mid \Cov_{1:t-1}, \Obs_{1:t-1}] &= \int \P(\Obs_{t+1} = \obs\mid\Cov_{t+1}=\cov, \Cov_{1:t},\Obs_{1:t})\cdot d(p(\Obs_t, \Cov_t\mid \Cov_{1:t-1},\Obs_{1:t-1}))\\
    &= \P(\Obs_{t+1}=\obs\mid\Cov_{t+1} = \cov, \Cov_{1:t-1},\Obs_{1:t-1})\\
    &= p_{t-1}(\obs\mid\cov).
\end{align*}
where the last equality follows from the c.i.d. assumption~\ref{def:cid}. We can then apply the martingale convergence theorem to obtain
\begin{align*}
    p_t(\obs\mid\cov) \to p_\infty(\obs\mid\cov) \quad \forall \obs,\cov, \mbox{a.s.}
\end{align*}
\section{Proof of Theorem~\ref{thm:perplexity-limit}}
\label{sec:proof-perplexity-limit}


  We begin by noting that under the cid condition~\eqref{eqn:cid}, the random
  measures $p_t$, $\widehat{p}_t$ are martingales for $\forall \obs$, under
  the filtration generated by $\Obs_{1:t}$ and $\what{\Obs}_{1:t}$
  respectively.  This is because
  \begin{equation}
    \E[p_t(\obs) \mid \Obs_{1:t-1}]
    = \E[ \indic{\Obs_{t+1} = \obs} \mid \Obs_{1:t-1}]
    = p_{t-1}(\obs),
  \end{equation}
  where the first equality is due to the tower law and the second equality
  follows from the c.i.d. condition. We can also apply this to the
  $\what{p}_t$ sequence. The martingale convergence theorem gives
  \[p_t(y) \to p_\infty(y),
    ~~\what{p}_t(y)\to \what{p}_\infty(y) ~~ \forall y, a.s.\]
  for some $p_{\infty}(y)$ and $\what{p}_{\infty}(y)$. Since the limit of
  measurable functions is measurable, we can show that the limiting quantities
  are valid random measures over $y$.
  By dominated convergence, we have
  \[\int p_t(y)\log \what{p}_t(y)dy\to\int p_\infty(y)\log\what{p}_\infty(y)dy\quad \mbox{a.s.}.\]

  To show the desired result, we use the decomposition
  \begin{align*}
    & \frac{1}{T} \sum_{t=1}^T \log
    \what{p}_t(\Obs_{t+1})- \int p_\infty(\obs) \log \what{p}_\infty(\obs) d\obs \\
    &=\frac{1}{T} \sum_{t=1}^T \biggl(\log
      \what{p}_t(\Obs_{t+1}) - \int p_t(\obs) \log \what{p}_t(\obs) d\obs\biggr)
      + \frac{1}{T}\sum_{t=1}^T\biggl(\int p_t(\obs) \log \what{p}_t(\obs) d\obs
      - \int p_\infty(\obs) \log \what{p}_\infty(\obs) d\obs\biggr).
  \end{align*}
  The second average converges to zero since individual elements converges to
  zero; it now suffices to show that the first average converges to zero.
  Evidently, the sequence
  \begin{equation*}
    D_t = \log \what{p}_t(\Obs_{t+1}\mid \Obs_{1:t})- \int p_t(\obs) \log
    \what{p}_t(\obs) d\obs = \log \what{p}_t(\Obs_{t+1}\mid \Obs_{1:t}) -
    \E[\log \what{p}_t(\Obs_{t+1})\mid \Obs_{1:t}]
  \end{equation*}
  is a martingale difference sequence adapted to the filtration $\mc{F}_t$
  generated by $Y_{1:t}$.  We use the following basic lemma due
  to~\citet[Corollary 2]{Teicher98}.
  \begin{lemma}[SLLN for martingale differences]
    \label{lemma:slln-martingale}
    Let $\{M_t\}$ be an $L^2$-martingale, and let $\{D_t\}$ be the corresponding
    martingale difference sequence. If $\E[D_t^2] < \infty$ and
    condition~\eqref{eqn:regularity} holds, then $M_t/t \to 0$ almost surely.
  \end{lemma}
  By Lemma~\ref{lemma:slln-martingale}, we have
  $\frac{1}{T}\sum_{t=1}^T D_t \to 0$ almost surely.

  ~\newline
  In the contextual case, we can follow the same proof by replacing the martingale differences with
  \begin{align*}
    D_t &= \log \what{p}_t(\Obs_{t+1}\mid \Cov_{t+1},\Obs_{1:t}, \Cov_{1:t})- \underset{\Cov_{t+1}\sim P_\Cov}{\E}\left[\underset{\Obs_{t+1}\sim p_t(\cdot\mid \Cov_{t+1})}{\E}[\log \what{p}_t(\Obs_{t+1}\mid \Cov_{t+1},\Obs_{1:t}, \Cov_{1:t})\mid \Cov_{t+1}]\right]\\
    &= \log \what{p}_t(\Obs_{t+1}\mid \Cov_{t+1})- \underset{\Cov\sim P_\Cov}{\E}\left[\int\log \what{p}_t(\obs\mid \Cov)p_t(\obs \mid \Cov)\right].
  \end{align*}
  Since we know that the covariates are drawn indpendently from $P_\Cov$, and we again omit the $\Obs_{1:t}, \Cov_{1:t}$ in the conditional when $p_t$ is indicitive of them.

  Then by the same argument, we have
  \begin{align*}
    & \frac{1}{T} \sum_{t=1}^T \log
    \what{p}_t(\Obs_{t+1}\mid \Cov_{t+1})- \underset{\Cov\sim P_\Cov}{\E}\int p_\infty(\obs\mid \Cov) \log \what{p}_\infty(\obs\mid \Cov) d\obs \\
    &=\frac{1}{T} \sum_{t=1}^T \biggl(\log
      \what{p}_t(\Obs_{t+1}\mid \Cov_{t+1}) - \underset{\Cov\sim P_\Cov}{\E}\int p_t(\obs\mid \Cov) \log \what{p}_t(\obs\mid \Cov) d\obs\biggr)\\
      &\quad + \frac{1}{T}\sum_{t=1}^T\biggl(\underset{\Cov\sim P_\Cov}{\E}\int p_t(\obs\mid \Cov) \log \what{p}_t(\obs\mid \Cov) d\obs
      - \underset{\Cov\sim P_\Cov}{\E}\int p_\infty(\obs\mid \Cov) \log \what{p}_\infty(\obs\mid \Cov) d\obs\biggr).
  \end{align*}
  The first term converges to zero by the same argument as before. For the second term, by the martingale convergence theorem, we have for any instantiation of $\cov,\obs$, we have
  \[p_t(\obs\mid\cov)\log\what{p}_t(\obs\mid \cov) \to p_\infty(\obs\mid\cov)\log\what{p}_\infty(\obs\mid \cov) \quad \mbox{a.s.}\]
  Then applying dominated convergence twice, we have that each term 
  \[\underset{\Cov\sim P_\Cov}{\E}\int p_t(\obs\mid\Cov)\log\what{p}_t(\obs\mid \Cov) d\obs \to \underset{\Cov\sim P_\Cov}{\E}\int p_\infty(\obs\mid\Cov)\log\what{p}_\infty(\obs\mid \Cov) d\obs\]
  Therefore we have that the second term converges to zero, and the proof is complete.


\section{Proof of Theorem~\ref{thm:regret}}
\label{sec:regret-proof}

Since $Y_{1:T}$ are i.i.d. under $Q$ and $\what{p}_{\theta}$, we have
 the following decomposition of the log likelihood ratio
\begin{align*}
  \E_{\Obs_{1:T}\sim Q}&\left[\frac{1}{T}[\log q(\Obs_{1:T}) - \log \what{p}(\Obs_{1:T})]\right]\\
  &= \E_{\Obs_{1:T}\sim Q}\left[\frac{1}{T} [\log q(\Obs_{1:T}) -\log \what{p}_{\theta\opt} (\Obs_{1:T})
    + \log \what{p}_{\theta\opt}(\Obs_{1:T}) - \log \what{p}(\Obs_{1:T})]\right]\\
  &= \dkl{Q}{\what{P}_{\theta\opt}}
    +\E_{\Obs_{1:T}\sim Q}\left[\frac{1}{T}[\log \what{p}_{\theta\opt}(\Obs_{1:T}) - \log \what{p}(\Obs_{1:T})]\right].
\end{align*}
Our proof for the convergence of
$\E_{\Obs_{1:T}\sim Q}\left[\log \what{p}_{\theta\opt}(\Obs_{1:T}) - \log
  \what{p}(\Obs_{1:T})\right]$ is inspired by that of~\citet{ClarkeBa90}, but
we make rexquisite modifications to properly bound error terms.

We prove the convergence for any instantiation $\obs_{1:T}$ of the random
variables $\Obs_{1:T}$. Following a similar notation as~\citet{ClarkeBa90},
define
\begin{equation*}
  B(\what{\theta},\delta) = \{\theta: \|\theta-\what{\theta}\|_{V_{\theta\opt}}\leq \delta\}
  \quad\text{ a neighborhood of } \what{\theta}~~\mbox{where}~~  \|\theta\|^2_{V_{\theta\opt}}
  \defeq \theta V_{\theta\opt}\theta^T
\end{equation*}
and the modulus of continuity of the log prior 
\[\rho(\delta,\theta\opt) = \sup_{\theta\in B(\theta\opt,\delta)}\left|\log\frac{\pi(\theta)}{\pi(\theta\opt)}\right|.\]
For fixed $K>d$, we will consider the radius $\delta_T = \sqrt{K/T}$. The
standardized score funciton is again defined by
\begin{equation*}
  S_T(\theta) := \frac{1}{\sqrt{T}}\sum_{t=1}^T
  \dot{\ell}_{\theta}(\Obs_t)
  ~~\mbox{where}~~
  \dot{\ell}_{\theta}(\obs)
  := \nabla_\theta \log \what{p}_\theta(\obs).
\end{equation*}
Define the moving estimator $\what{\theta}$
\begin{equation}
    \label{eq:theta-hat}
    \what{\theta} = \theta\opt + \frac{1}{\sqrt{T}}V_{\theta\opt}^{-1}S_T
    \indic{S_T^TV_{\theta\opt}^{-1}S_T \leq K}
\end{equation}
where for notational simplicity, we omit the dependence on $\theta\opt$ in the
$S_T$, i.e., $S_T \defeq S_T(\theta\opt)$.  Note that by definition,
$\norm{\what{\theta} - \theta\opt}_{V_{\theta\opt}} \le \delta_T$.

\paragraph{Controlling the log likelihood difference by restricting to the neighborhood $B(\what{\theta}, \delta_T)$}
 In the following lemma, we bound the difference between the log likelihoods
by focusing on a neighborhood of $\what{\theta}$ where the posterior is
concentrated. We defer its proof to
Section~\ref{section:proof-log-like-bound}.
\begin{lemma}
  \label{lemma:log-like-bound}
  Denote the truncated normal centered at $\what{\theta}$ as
  \begin{equation}
    \label{eqn:pdf}
  \phi_T(\theta) = \frac{1}{c_T}
  \exp\left(- \frac{T}{2}\|\theta-\what{\theta}\|_{V_{\theta\opt}}^2\right)
  \indic{\theta\in B(\what{\theta},\delta_T)}
\end{equation}
where $c_T \defeq
\int_{B(\what{\theta},\delta_T)}e^{-(T/2)\|\theta-\what{\theta}\|_{V_{\theta\opt}}^2}d\theta$.
Then, we have the bound
\begin{align}
  \log \what{p}_{\theta\opt}(\obs_{1:T})
  - \log \what{p}(\obs_{1:T})
  & \le  \log \frac{1}{c_T} + \rho(2\delta_T, \theta\opt)+ \frac{1}{2}S_T^T V_{\theta\opt}^{-1}S_T \nonumber \\
  & \qquad + \int_{B(\what{\theta},\delta_T)}
    \left(\log \frac{\what{p}_{\theta\opt}(\obs_{1:T})}
    {\what{p}_\theta(\obs_{1:T})}
    - \frac{T}{2}\|\theta-\theta\opt\|_{V_{\theta\opt}}^2
    - \sqrt{T}(\theta\opt - \theta)^T S_T\right) \phi_T(\theta)d\theta \nonumber \\
  & \qquad + \frac{3}{2}S_T^T V_{\theta\opt}^{-1}S_T
    \indic{S_T^T V_{\theta\opt}^{-1}S_T>K} + \log \frac{1}{\pi(\theta\opt)}.
    \label{eq:log-ineq2}
\end{align}
\end{lemma}

We proceed by bounding each term in the inequality~\eqref{eq:log-ineq2}.

\paragraph{Bounding the leading term $\log \frac{1}{c_T}$}
We can upper bound $c_T$ using the normalizing constant of the normal
distribution
\[c_T \le (2\pi)^{d/2}T^{-d/2} \det V_{\theta\opt}^{-1/2}.\] The lower bound
follows from Chebyshev's inequality. Consider $\theta$ as a Gaussian random
variable centered around $\widehat{\theta}$ with covariance
$(TV_{\theta^\star})^{-1}$. Then by Chebyshev's inequality,
\begin{align*}
  \int_{B(\what{\theta},\delta_T)}e^{-(T/2)\|\theta-\what{\theta}\|_{V_{\theta^\star}}^2}d\theta
  &= ((2\pi)^{d/2}T^{-d/2}\det V_{\theta^\star}^{-1/2})
    \cdot\P(\theta\in B(\what{\theta},\delta_T))\\
  &= ((2\pi)^{d/2}T^{-d/2}\det V_{\theta^\star}^{-1/2})
    (1- \P(\|\theta-\what{\theta}\|_{V_{\theta^\star}}\geq \delta_T))\\
  &\geq ((2\pi)^{d/2}T^{-d/2}\det V_{\theta^\star}^{-1/2})\cdot (1- d/K)
\end{align*}
Therefore, we have that
\[(1-d/K)(2\pi)^{d/2}T^{-d/2} \det V_{\theta^\star}^{-1/2}
  \le c_T \le (2\pi)^{d/2}T^{-d/2} \det V_{\theta^\star}^{-1/2},\]
or equivalently,
\begin{equation}
  \label{eq:const-approx}
  0 \le \log \frac{1}{c_T} - \frac{d}{2} \log \frac{T}{2\pi}
  - \frac{1}{2} \log\det V_{\theta\opt}
  \le \log \frac{K}{K-d}.
\end{equation}

\paragraph{Controlling the pdf $\phi_T(\theta)$~\eqref{eqn:pdf}}
To bound the integral term in the inequality~\eqref{eq:log-ineq2}, note that
for any $\theta\in B(\what{\theta},\delta_T)$, we have that
\begin{align*}
  \|\theta-\theta\opt\|_{V_{\theta\opt}}^2
  \le \left( \norm{\theta - \what{\theta}}_{V_{\theta\opt}}
  + \norm{\what{\theta} - \theta\opt}_{V_{\theta\opt}}\right)^2
  \le 
  \|\theta-\what{\theta}\|_{V_{\theta\opt}}^2 + 4\delta_T^2 =
  \|\theta-\what{\theta}\|_{V_{\theta\opt}}^2 +4\frac{K}{T}.
\end{align*}
We can then bound
the density $\phi_T(\theta)$ which is centered around $\what{\theta}$ with the
above density centered around $\theta\opt$ by
\begin{align*}
  \phi_T(\phi) &= 1/c_T\cdot
  \exp\left(-\frac{T}{2}\|\theta-\what{\theta}\|_{V_{\theta\opt}}^2\right)
  \indic{\theta\in B(\what{\theta},\delta_T)} \\
  & \leq 1/c_T \cdot e^{2K}\cdot
  \exp\left(-\frac{T}{2}\|\theta-\theta\opt\|_{V_{\theta\opt}}^2\right)
  \indic{\theta\in B(\theta\opt,2\delta)} \\
  & \leq (1-d/K)^{-1}e^{2K}\phi^\star_T(\theta)
\end{align*}
where the density
$ \phi^\star_T = \mathcal{N}(\theta\opt, (TV_{\theta\opt})^{-1})$, and the
last inequality follows from the lower bound of $c_T$.
Hence, the integral term in the inequality~\eqref{eq:log-ineq2} can be bounded by
\begin{equation}
  \label{eq:pdf-bound}
  \frac{e^{2K}}{1-d/K}
  \int_{B(\theta\opt,2\delta_T)}\phi_T^\star(\theta)\left|\log
    \frac{\what{p}_{\theta\opt}(\obs_{1:T})}{\what{p}_\theta(\obs_{1:T})} -
    \frac{T}{2}\|\theta-\theta\opt\|_{V_{\theta\opt}}^2 - \sqrt{T}(\theta\opt
    - \theta)^T S_T\right|d\theta.
\end{equation}

\paragraph{Controlling the integrand in the bound~\eqref{eq:log-ineq2} via Taylor expansion}
Noting that $\obs_{1:T}$ are i.i.d. under $\what{p}_{\theta}$,
we can also break down the integrand
\begin{align*}
  & \left|\log \frac{\what{p}_{\theta\opt}(\obs_{1:T})}
    {\what{p}_\theta(\obs_{1:T})}
    - \sqrt{T}(\theta\opt - \theta)^T S_T- \frac{T}{2}
    \norm{\theta-\theta\opt}_{V_{\theta\opt}}^2 \right| \\
  & = \left| \sum_{t=1}^T \log \frac{\what{p}_{\theta\opt}(\obs_{t})}
    {\what{p}_\theta(\obs_{t})}
    - \sqrt{T}(\theta\opt - \theta)^T S_T- \frac{T}{2} 
    \norm{\theta-\theta\opt}_{V_{\theta\opt}}^2 \right| \\
  & \leq  \left|
    \sum_{t=1}^T
    \log \frac{\what{p}_{\theta\opt}(\obs_{t})}{\what{p}_\theta(\obs_{t})}
         - (\theta\opt - \theta)^T \dot{\ell}_{\theta^\star}(\obs_t)
         - \frac{1}{2}(\theta\opt-\theta)^T \nabla^2\log \widehat{p}_{\theta\opt}(\obs_t)
         (\theta\opt-\theta)\right| \\
  &\qquad + \frac{1}{2}\left|(\theta\opt-\theta)^T
    \sum_{t=1}^T \left( \nabla^2\log \widehat{p}_{\theta\opt}(\obs_t) - V_{\theta\opt}\right)
    (\theta\opt-\theta)\right|.
\end{align*}
The first term in the preceding bound can be directly controlled via a
second order Taylor expansion of the log likelihood
\begin{equation*}
  \left|\log \frac{\what{p}_{\theta\opt}(\obs_t)}{\what{p}_\theta(\obs_t)}  - (\theta\opt - \theta)^T \dot{\ell}_{\theta\opt}(\obs_t) - \frac{1}{2}(\theta\opt-\theta)^T \nabla^2\log \widehat{p}_{\theta\opt}(\obs_t)(\theta\opt-\theta)\right|
   = o(\|\theta-\theta\opt\|_{V_{\theta\opt}}^2).
\end{equation*}
To bound the second term, note that there exists a universal constant $c> 0$ such that
\begin{align*}
  \left|(\theta\opt-\theta)^T
    \left(\frac{1}{T} \sum_{t=1}^T \nabla^2
    \log \widehat{p}_{\theta\opt}(\obs_t) - V_{\theta\opt}\right)
    (\theta\opt-\theta)\right|
   & \leq \|\theta\opt-\theta\|^2_2\cdot
    \left\|\frac{1}{T}\sum_{t=1}^T
    \nabla^2\log \widehat{p}_{\theta\opt}(\obs_t) - V_{\theta\opt}\right\|_2 \\
  & \le c\cdot d \cdot \|\theta\opt-\theta\|_{V_{\theta\opt}}^2 \cdot
    \left\|\frac{1}{T}\sum_{t=1}^T
    \nabla^2\log \widehat{p}_{\theta\opt}(\obs_t) - V_{\theta\opt}\right\|_2 \\
   & \leq c\cdot d\cdot \delta_T^2 \left\|\frac{1}{T}\sum_{t=1}^T
    \nabla^2\log \widehat{p}_{\theta\opt}(\obs_t) - V_{\theta\opt}\right\|_2.
\end{align*}
Collecting these approximations and plugging them into our previous
bound~\eqref{eq:pdf-bound}, we arrive at
\begin{align}
  \label{eq:bound-integral}
  & \int_{B(\what{\theta},\delta_T)}
    \left(\log \frac{\what{p}_{\theta\opt}(\obs_{1:T})}
    {\what{p}_\theta(\obs_{1:T})}
    - \frac{T}{2}\|\theta-\theta\opt\|_{V_{\theta\opt}}^2
  - \sqrt{T}(\theta\opt - \theta)^T S_T\right) \phi_T(\theta)d\theta \\
  & \le 
    \frac{e^{2K}}{1-d/K}\left(
    d\cdot  o(1) + \frac{cdK}{2} \cdot \norm{
    \frac{1}{T}\sum_{t=1}^T\nabla^2\log \widehat{p}_{\theta\opt}(\obs_t) - V_{\theta\opt}}_2 \right).
\end{align}

\paragraph{Setting $K$ large to control the final approximation error}
Returning to the inequality~\eqref{eq:log-ineq2} in
Lemma~\ref{lemma:log-like-bound}, consider the approximation error 
\begin{align*}
  \mathfrak{R}_T
  \defeq \log\what{p}_{\theta\opt}(\obs_{1:T})
    - \log \what{p}(\obs_{1:T})
  - \left(\frac{d}{2}\log\frac{T}{2\pi} + \log\frac{1}{\pi(\theta\opt)} + \frac{1}{2}\log \det(V_{\theta\opt})
  - \frac{1}{2}S_T^T V_{\theta\opt}^{-1}S_T\right).
\end{align*}
So far, we have shown
\begin{align*}
  \mathfrak{R}_T
  & \le   \rho\left( 2 \sqrt{\frac{K}{T}},\theta\opt\right)
  + \frac{Ke^{2K}}{K-d}
  \left(d \cdot o(1) + dK \norm{
  \frac{1}{T}\sum_{t=1}^T\nabla^2\log \widehat{p}_{\theta\opt}(\obs_t) - V_{\theta\opt}}_2
  \right) \\
  & \qquad + \frac{3}{2} S_T^\top V_{\theta\opt}^{-1} S_T \indic{S_T^\top V_{\theta\opt}^{-1} S_T > K}.
\end{align*}
Fix an arbitrarily small $\epsilon > 0$.  Since
$S_T^\top V_{\theta\opt}^{-1} S_T = O(1)$ by the strong law of large numbers,
we can pick $K$ large enough such that there exists $T_0$ such that for all
$T \ge T_0$,
$\frac{3}{2} S_T^\top V_{\theta\opt}^{-1} S_T \indic{S_T^\top
  V_{\theta\opt}^{-1} S_T > K} \le \epsilon$.  For such choice of $K$, uniform
positivity and continuity of $\pi(\cdot)$ guarantees that
\[
  \rho\left( 2 \sqrt{\frac{K}{T}},\theta\opt\right) \to 0
  ~~\mbox{as}~~T \to \infty.
\]
This gives the desired result.

\subsection{Proof of Lemma~\ref{lemma:log-like-bound}}
\label{section:proof-log-like-bound}

We begin by rewriting the log likelihood ratio as
\begin{align*}
  \log\frac{\what{p}_{\theta\opt}(\obs_{1:T})\pi(\theta\opt)}{\what{p}(\obs_{1:T})} =
  & -\log\int \frac{\what{p}_\theta(\obs_{1:T})\pi(\theta)}{\what{p}_{\theta\opt}(\obs_{1:T})\pi(\theta\opt)}d\theta\\
  \leq
  &-\log\int_{B(\what{\theta},{\delta_T})} \frac{\what{p}_\theta(\obs_{1:T})\pi(\theta)}{\what{p}_{\theta\opt}(\obs_{1:T})\pi(\theta\opt)}d\theta\\
  = & -\log\int_{B(\what{\theta},{\delta_T})}
      \frac{c_T\cdot\what{p}_\theta(\obs_{1:T})\exp(T/2 \cdot \|\theta-\what{\theta}\|_{V_{\theta\opt}}^2) \pi(\theta)}
      {\what{p}_{\theta\opt}(\obs_{1:T})\pi(\theta\opt)}\phi_T(\theta) d\theta.
\end{align*}
We use Jensen's inequality to put the log inside the integral
\begin{align}
    \label{eq:log-ineq}
  \log\frac{\what{p}_{\theta\opt}(\obs_{1:T})\pi(\theta\opt)}{\what{p}(\obs_{1:T})} \leq
  &  -\log\int_{B(\what{\theta},{\delta_T})}
    \frac{c_T\cdot\what{p}_\theta(\obs_{1:T})\exp(T/2 \cdot
    \|\theta-\what{\theta}\|_{V_{\theta\opt}}^2)
    \pi(\theta)}{\what{p}_{\theta\opt}(\obs_{1:T})
    \pi(\theta\opt)}\phi_T(\theta) d\theta \\
  \leq & -\log c_T
         + \sup_{\theta\in B(\widehat{\theta},\delta_T)}
         \left|\log \frac{\pi(\theta)}{\pi(\theta\opt)}\right|
         + \int_{B(\widehat{\theta},\delta_T)}
         \left(\log \frac{\what{p}_{\theta\opt}(\obs_{1:T})}
         {\what{p}_\theta(\obs_{1:T})}
         - \frac{T}{2}
         \|\theta-\what{\theta}\|_{V_{\theta\opt}}^2\right)
         \phi_T(\theta)d\theta \nonumber \\
  \leq & -\log c_T + \rho(2\delta_T, \theta\opt)
         +\int_{B(\widehat{\theta},\delta_T)}
         \left(\log \frac{\what{p}_{\theta\opt}(\obs_{1:T})}
         {\what{p}_\theta(\obs_{1:T})}
         - \frac{T}{2}\|\theta-\what{\theta}\|_{V_{\theta\opt}}^2\right)
         \phi_T(\theta)d\theta \nonumber 
\end{align}
where the last inequality follows form triangle inequality
$\|\theta-\theta\opt\|_{V_{\theta\opt}}\leq
\|\theta-\widehat{\theta}\|_{V_{\theta\opt}}+\|\what{\theta}-\theta\opt\|_{V_{\theta\opt}}\leq
2\delta_T$.

Using the definition of $\what{\theta}$~\eqref{eq:theta-hat}, we can rewrite
$\|\theta - \what{\theta}\|_{V_{\theta\opt}}^2$
\begin{align*}
  \frac{T}{2} \|\what{\theta}-\theta\|_{V_{\theta\opt}}^2 =
  & \frac{T}{2}\norm{\theta - \theta\opt
    - \frac{1}{\sqrt{T}} V_{\theta\opt}^{-1}S_T
    \indic{S_T^TV_{\theta\opt}^{-1}S_T\leq K}}_{V_{\theta\opt}}^2\\
    =& \frac{T}{2}\|\theta-\theta\opt\|_{V_{\theta\opt}}^2 + \sqrt{T}(\theta\opt-\theta)^T S_T + \frac{1}{2}S_T^T V_{\theta\opt}^{-1}S_T \\
    &\qquad - \left(\sqrt{T}(\theta\opt-\theta)^T S_T + \frac{1}{2}S_T^T V_{\theta\opt}^{-1}S_T \right)\mathbbm{1}_{S_T^T V_{\theta\opt}^{-1}S_T>K}
\end{align*}
by simply expanding the norm term.

Then by Cauchy-Schwarz, under the event $S_T^T V_{\theta\opt}^{-1}S_T>K$ and $\theta \in B(\what{\theta},\delta_T)$
\begin{align*}
    \sqrt{T}|(\theta - \what{\theta})^TS_T|&\leq \sqrt{T}\|\theta-\what{\theta}\|_{V_{\theta\opt}}\cdot\|S_T\|_{V_{\theta\opt}^{-1}}\leq\sqrt{T}\delta_T(S_T^T V_{\theta\opt}^{-1}S_T)^{1/2}\\
    & = \sqrt{K}(S_T^T V_{\theta\opt}^{-1}S_T)^{1/2}\leq S_T^T V_{\theta\opt}^{-1}S_T
\end{align*}
Putting this together, we have the inequality
\[\frac{T}{2}\|\theta-\what{\theta}\|_{V_{\theta\opt}}^2 \geq \frac{T}{2}\|\theta-\theta\opt\|_{V_{\theta\opt}}^2 +\sqrt{T}(\theta\opt-\theta)^T S_T + \frac{1}{2}S_T^T V_{\theta\opt}^{-1}S_T - \frac{3}{2}S_T^T V_{\theta\opt}^{-1}S_T\mathbbm{1}(S_T^T V_{\theta\opt}^{-1}S_T>K)\]
Substituting this inequality into the previous bound on the likelihood
ratio~\eqref{eq:log-ineq}, conclude
\begin{align*}
    \log&\frac{\what{p}_{\theta\opt}(\obs_{1:T})\pi(\theta\opt)}{\what{p}(\obs_{1:T})}\\
    \leq &-\log c_T + \rho(2\delta_T, \theta\opt)+ \frac{1}{2}S_T^T V_{\theta\opt}^{-1}S_T \\
    &+\int_{B(\what{\theta},\delta_T)} \left(\log \frac{\what{p}_{\theta\opt}(\obs_{1:T})}{\what{p}_\theta(\obs_{1:T})} - \frac{T}{2}\|\theta-\theta\opt\|_{V_{\theta\opt}}^2 - \sqrt{T}(\theta\opt - \theta)^T S_T\right) \phi_T(\theta)d\theta\\
    & + \frac{3}{2}S_T^T V_{\theta\opt}^{-1}S_T\mathbbm{1}(S_T^T V_{\theta\opt}^{-1}S_T>K).
\end{align*}


\section{One-layer Transformer for Bayesian Linear Regression}
\label{section:transformers}

\subsection{Proof of Lemma~\ref{lemma:gd}}
\label{proof:gd}
We replicate the proof of~\citet{WuZoChBrGuBa23} for completeness. Recalling
the input matrix $\mathbf{Z}_{t}$~\eqref{eqn:io}, we drop the subscript to
ease notation. By definition, the output of the self-attention layer is given
by
\begin{align*}
  \what{\obs}_t &= \left[ \mathbf{Z} + \frac{1}{n} (\mathbf{V}\mathbf{Z})(\mathbf{QZ})^\top (\mathbf{KZ}) \right]_{d+1,t} \\
               &= \mathbf{e}_{d+1}^\top \left( \mathbf{Z} + \frac{1}{n} \mathbf{V}\mathbf{Z}\mathbf{Z}^\top \mathbf{Q}^\top \mathbf{KZ} \right) \mathbf{e}_{t} \\
               &= 0 + \frac{1}{t-1} \mathbf{e}_{d+1}^\top \mathbf{V}\mathbf{Z}\mathbf{Z}^\top \mathbf{Q}^\top \mathbf{KZ}\mathbf{e}_{t} \\
               &= \frac{1}{t-1} (\mathbf{e}_{d+1}^\top \mathbf{V}) \left( \begin{bmatrix} \vec{\Cov}_{t-1}^\top \vec{\Cov}_{t-1} + \cov_t\cov^t_t & \vec{\Cov}_{t-1}^\top \vec{\Obs}_{t-1} \\ \vec{\Obs}_{t-1}^\top \vec{\Cov}_{t-1} & \vec{\Obs}_{t-1}^\top \vec{\Obs}_{t-1} \end{bmatrix}  \right) \mathbf{Q}^\top \mathbf{K} \begin{bmatrix} \cov_t \\ 0 \end{bmatrix}
\end{align*}
The key assumption is that the bottom left $1\times d$ blocks of the $V$ and
$Q^TK$ matrices are fixed $0$, that is
\begin{align*}
  \mathbf{V} = \begin{bmatrix}
    * & *\\
    0 & v
  \end{bmatrix}
        \quad \text{and} \quad
        \mathbf{Q^TK} = \begin{bmatrix}
          \mathbf{W} & *\\
          0 & *
        \end{bmatrix}
\end{align*}
where $v$ is a scalar and $\mathbf{W}$ is a $d\times d$ matrix. This implies
that the output of the transformer is
\[
  \what{\obs}_t = \frac{1}{t}
  \vec{\Obs}_{t-1}^\top \vec{\Cov}_{t-1} \mathbf{W} v^\top
  \cov_t.
\]
Letting $\Gamma^\top \defeq \mathbf{W} v^\top$, we have the desired reparameterization.




\subsection{Proof of Proposition~\ref{prop:blr-main-term-convergence}}
\label{proof:example-blr-main}

We start with the usual chain rule for KL divergences
\begin{align*}
  & \frac{1}{T}\E_Q[\log q(\Obs_{1:T}\mid \Cov_{1:T})
  -\log\what{p}(\Obs_{1:T}\mid \Cov_{1:T})] \\
  & = \frac{1}{T}\sum_{t=1}^T \E_Q\E_Q\left[
  \log \frac{q(\Obs_t \mid \Cov_{1:t}, \Obs_{1:t-1})}
  {\what{p}(\Obs_t \mid \Cov_{1:t}, \Obs_{1:t-1})}
  \mid \Cov_{1:t}, \Obs_{1:t-1}
  \right] \\
  & = \frac{1}{T}\sum_{t=1}^T \E_Q
    \left[\dkl{N(w_q^\top \Cov_t, \sigma^2)}
    {N(\what{\mu}_t(\Cov_t), \what{\sigma}_t^2)}\right] \\
  &=\frac{1}{2T} \sum_{t=1}^T \E_Q \left[
      \left(\frac{\sigma^2}{\what{\sigma}_t^2}-1\right)
      + \log\frac{\what{\sigma}_t^2}{\sigma^2}
      + \frac{1}{\what{\sigma}_t^2}
      \left(w_q^\top \Cov_t - \frac{1}{t} \vec{\Obs}_{t-1}^\top \vec{\Cov}_{t-1} \Gamma^\top \Cov_t
      \right)^2
      \right].
\end{align*}
In the final equality, we used the formula
$\dkl{N(\mu_1, v_1^2)}{N(\mu_2, v_2^2)} = \log\frac{v_2}{v_1} + \frac{1}{2
  v_2^2} (v_1^2 - v_2^2)+ \frac{1}{2 v_2^2} (\mu_1 - \mu_2)^2$.  Recall the
autoregressive mean~\eqref{eqn:reparam} and the
definition~\eqref{eqn:posterior-var} of the autoregressive (posterior)
variance $\what{\sigma}_t = \sigma^2+\Cov_t^TA_{t-1}^{-1}\Cov_t$.  Since
$\Cov_t$ is independent of $A_{t-1}, \vec{\Obs}_{t-1}, \vec{\Cov}_{t-1}$,
conclude
\begin{align*}
  & \frac{1}{T}
  \E_Q[\log q(\Obs_{1:T}\mid X_{1:T})
    -\log\what{p}(\Obs_{1:T}\mid X_{1:T})] \\
  & = \frac{1}{2T} \sum_{t=1}^T \E_Q \left[
    \frac{-\Cov^\top A_{t-1} \Cov}{\sigma^2 + \Cov^\top A_{t-1} \Cov}
    + \log\left( 1 + \frac{\Cov^\top A_{t-1} \Cov}{\sigma^2}\right)
    + \frac{1}{\sigma^2 + \Cov^\top A_{t-1} \Cov}
    \left(w_q^\top \Cov - \frac{1}{t} \vec{\Obs}_{t-1}^\top \vec{\Cov}_{t-1} \Gamma^\top \Cov
    \right)^2
    \right].
\end{align*}

To show the limit, note that since $A_{t-1}$ is positive semidefinite, the
integrand in the preceding display (which is clearly positive) is bounded above by
\begin{equation*}
  0
  + \frac{\Cov^\top A_{t-1} \Cov}{\sigma^2}
  + \frac{1}{\sigma^2}
  \left(w_q^\top \Cov - \frac{1}{t} \vec{\Obs}_{t-1}^\top \vec{\Cov}_{t-1} \Gamma^\top \Cov
  \right)^2,
\end{equation*}
which is evidently integrable. Applying dominated convergence and noting that
$\Cov^\top A_{t-1} \Cov \cas 0$ and
$\frac{1}{t} \vec{\Obs}_{t-1}^\top \vec{\Cov}_{t-1} \cas \E_Q[\Obs \Cov^\top] =
\E_Q[(w_q^\top \Cov + \varepsilon) \Cov^\top] = w_q^\top H$, we have shown a stronger result that each summand converges to
\begin{equation*}
  \sigma^{-2}
  \E\left[
    \left(w_q^\top \Cov - w_q^\top H \Gamma^\top \Cov
  \right)^2
    \right] = \sigma^{-2} w_q^\top (I - H \Gamma^\top) H (I - H \Gamma^\top)^\top w_q.
\end{equation*}


\section{Regularization}
\label{section:regularization}

Recall we want to "enforce" the exchangeability/cid condition~\eqref{eqn:cid}
on our fitted model by adding a penalty term to the loss function, as a form
of regularization. Specifically, we could consider regularizing with the KL
divergence between the one- vs. two-step forward
prediction~\eqref{eqn:kl-regularized}. In the contextual setting with realizations $\cov_{1:T}^i$ as covariates and $\obs_{1:T}^i$ as observables, the regularized training objective for the $i$-th sequence is
\begin{align*}
 &\sum_{t=0}^{T-1}  \underbrace{\log \what{p}_t( \Obs_{t+1}^i = \obs_{t+1}^i | \Cov_{1:t}^i = \cov_{1:t}^i, \Obs_{1:t}^i = \obs_{1:t}^i, \Cov_{t+1}^i = \cov_{t+1}^i)}_{\text{autoregressive loss terms}}\\
  &+ \lambda \cdot \sum_{t=0}^{T-2} \dkl{\underbrace{\what{p}_t(\obs_{t+1}^i | \cov_{1:t}^i, \obs_{1:t}^i, \cov_{t+1}^i)}_{\begin{array}{c}\text{one-step forward pred}\\\text{ (same as autoreg loss terms)}\end{array}}}{\underbrace{
    \what{p}(\Obs_{t+2}^i = \obs_{t+1}^i \mid \Cov_{1:t}^i = \cov_{1:t}^i, \Obs_{1:t}^i = \obs_{1:t}^i, \Cov_{t+2}^i = \cov_{t+1}^i)
  }_{\begin{array}{c}
  \text{two-step forward pred} \\
  \text{(if } \obs_{t+1}^i, \cov_{t+1}^i \text{ had been observed two steps forward)}
  \end{array}}}
\end{align*}
where the $\lambda$ is a hyperparameter for the KL term.

Since the term $\what{p}(\Obs_{t+2}^i = \obs_{t+1}^i \mid \Cov_{1:t}^i = \cov_{1:t}^i, \Obs_{1:t}^i = \obs_{1:t}^i, \Cov_{t+2}^i = \cov_{t+1}^i)$
is not directly obtainable from an autoregressive sequence model, we estimate
it in the following derivation. 

Assuming that this term is a normal
distribution, the KL divergence between two normal distributions is
$\dkl{\mathcal{N}(\mu_1, \sigma^2_1)}{\mathcal{N}(\mu_2,\sigma^2_2)} =
\frac{1}{2} \left(\sigma_2^{-1}\sigma_1 -1 + \sigma_2^{-1}(\mu_2-\mu_1)^2 +\ln
  \frac{\sigma_2}{\sigma_1}\right)$. First,  we take a Monte Carlo estimate of the
two-step conditional mean. Again, for realizations $\cov_{1:T}^i$ as covariates and $\obs_{1:T}^i$ as observables, using the tower property
\begin{align*}
  \what{p}(\Obs_{t+2} = \obs\mid \cov_{1:t}, \obs_{1:t}, \Cov_{t+2} = x)
  = \int\int &\what{p}_{t+2}(\obs\mid \cov_{1:t}, \obs_{1:t}, \cov_{t+1}, \Obs_{t+1} = \zeta, \Cov_{t+2} =\cov) \\
  & \cdot \what{p}_{t+1}(\zeta\mid \cov_{1:t}, \obs_{1:t}, \Cov_{t+1} = \cov_{t+1})
               \cdot p_{\Cov}(\cov_{t+1}) 
    d\zeta d\cov_{t+1},
\end{align*}
Based on the above equatoin, rewrite the two-step conditional mean as
\begin{alignat*}{2}
\what{\mathbb{E}} \left[\Obs_{t+2}\mid \cov_{1:t}, \obs_{1:t}, \Cov_{t+2} = x\right] &= \int \int && \left(\int \phi\left(\mu_{t+2}(\cov_{t+1},\zeta,x), \sigma^2_{t+2}(\cov_{t+1}, \zeta,x)\right)(y)\cdot y dy\right) \\
& &&\cdot \phi\left(\mu_{t+1}(\cov_{t+1}),\sigma^2_{t+1}(\cov_{t+1})\right)(\zeta)\cdot \phi(\cov_{t+1}) d\cov_{t+1}d\zeta\\
&= \int\int &&\mu_{t+2}(\cov_{t+1}, \zeta, x) \cdot\phi\left(\mu_{t+1}(\cov_{t+1}),\sigma^2_{t+1}(\cov_{t+1})\right)(\zeta) \\
& &&\cdot\phi(\cov_{t+1}) d\zeta d\cov_{t+1}\\
&= \int\int &&\mu_{t+2}(\cov_{t+1}, \zeta, x) \cdot\what{p}_{t+1}(\zeta\mid \Cov_{1:t}, \Obs_{1:t}, \Cov_{t+1} = \cov_{t+1})\\
& &&\cdot P_{\Cov_{t+1}}(\cov_{t+1}) d\zeta d\cov_{t+1}\\
&= \underset{\Cov_{t+1}}{\E}
&&\underset{\zeta\sim \what{p}_{t+1}(|\Cov_{t+1})}{\E}
[ \mu_{t+2}(\Cov_{t+1}, \zeta, x)]
\end{alignat*}
where $\phi$ denotes the density of the normal distribution, and the equality follows from interchangebale intergrals. From the above equation, we can draw
$\Cov_{t+1}$ samples from a fixed covariate distribution $P_X$, and then draw
$\zeta$ samples given this $\cov_{t+1}$ from the distribution predicted by
model. 

Similarly, we can approximate the two-step conditional variance
\begin{alignat*}{2}
  \what{\mathbb{E}} \left[\Obs_{t+2}^2\mid \Cov_{1:t}, \Obs_{1:t}, \Cov_{t+2} = x\right] &= \int \int && \left(\int \phi\left(\mu_{t+2}(\cov_{t+1},\zeta,x), \sigma^2_{t+2}(\cov_{t+1}, \zeta,x)\right)(y)\cdot y^2 dy\right) \\
  & &&\cdot \phi\left(\mu_{t+1}(\cov_{t+1}),\sigma^2_{t+1}(\cov_{t+1})\right)(\zeta)\cdot \phi(\cov_{t+1}) d\cov_{t+1}d\zeta\\
  &= \int\int &&\left(\mu_{t+2}^2(\cov_{t+1}, \zeta, x) + \sigma^2_{t+2}(\cov_{t+1}, \zeta, x)\right)\\
  & &&\cdot\phi\left(\mu_{t+1}(\cov_{t+1}),\sigma^2_{t+1}(\cov_{t+1})\right)(\zeta)\cdot \phi(\cov_{t+1}) d\zeta d\cov_{t+1}\\
  &= \underset{\Cov_{t+1}}{\E}&&\underset{\zeta\sim \what{p}_{t+1}(|\Cov_{t+1})}{\E}[ \mu^2_{t+2}(\Cov_{t+1}, \zeta, x)+\sigma^2_{t+2}(\Cov_{t+1}, \zeta, x)]
  \end{alignat*}


\section{Experiments Details}
\label{section:experiments-details}

\subsection*{Model Architectures}

\subsubsection*{GPT2}

\begin{itemize}[noitemsep]
    \item Model dimension: 1
    \item Number of embeddings layers: 4
    \item Feed forward dimension: 128
    \item Number of attention heads: 4
    \item Number of transformer layers: 12
    \item Batch size: 32
    \item Number of training steps: 30000
    \item Learning rate: $1e^{-4}$ with Cosine annealing scheduler
\end{itemize}

For the transformers without positional embedding, we modified the Huggingface Transformer's GPT2 architecture to remove positional embeddings by setting them to \([0,0,0,\ldots,0]\). Additionally, we experimented with different positional embeddings \([sin, 0101]\), but found the results inconclusive and not critical to our theory.
The GPT2 architecture comprises 9 million parameters, whereas the Exchangeable Transformer has 220 thousand parameters.
We also experimented across different number of layers [ 6, 12 ] and heads for GPT2.

\subsubsection*{Exchangeable Transformer}

\begin{itemize}[noitemsep]
    \item Model dimension: 1
    \item Number of embeddings layers: 4
    \item Feed forward dimension: 128
    \item Number of attention heads: 4
    \item Number of transformer layers: 12
    \item Batch size: 32
    \item Number of training steps: 30000
    \item Learning rate: $1e^{-4}$ with Cosine annealing scheduler
\end{itemize}

\subsubsection*{TNP, PFN, and Exchangeable Transformer}
We construct each token through concatenating $(x_i,y_i)$ to preserve the pair wise structure of feature-label pairs. As in TNP, we also make use of auxiliary tokens consisting of  $(x_i,0)$. Not only does doing so remove the y label from the points that the model will learn to predict, but having such tokens - when combined with the attention scheme illustrated in Fig. 5 - preserves the autoregressive structure of our tasks.
Deviating from Nguyen et al’s work, we remove the context points which were allowed to attend to each other in TNP. We do so as our aim is to investigate autoregressive loss, which begins at predicting the 1st label while conditioning on no previous context. Removing the initial context also allows for a full examination of our model’s ability to encapsulate a Bayesianprior, as the prior is more pronounced than the likelihood when a Bayesianstatistician makes predictions based on a few or no context.
We further extend our contribution by augmenting TNP’s attention mechanism. We adjust the attention mask such that each auxiliary/padded token attends to itself. We introduce this improvement as this adaptation enables the model to access query value pairs associated with the current token’s index, whereas previously the model was only allowed to query information at previous indices. We observe that this improvement in informational access offers performance improvements, especially in short/zero context predictions. We posit that allowing for self attention helps the model embody knowledge for predicting each point, rather than forming predictions solely based on context.

\subsection*{Training Compute}
Training was conducted on 8x A100 GPUs. The CID-Regularizer, with its Monte Carlo Sampling for KL Divergence computation, necessitated parallel computation. All code was implemented in PyTorch, with data generated from deterministic and random seeds to average results across trajectories.

\subsection*{Parameter Inference}
For each batch, we draw one batch of \(X\) from a standard normal distribution for Bayesian Linear Regression and a uniform distribution \([-2, 2]\) for Gaussian Process to ensure stability. This \(X\) is then passed into the BLR or Gaussian Process function, with each batch of sequential data drawn from the same function/coefficient.

\subsection*{CID Regularizer}
Monte Carlo Sampling was used to compute the KL Divergence for CID Regularizer. We experimented with \(\lambda\) values \([0.001, 0.1, 1, 10, 100]\) and Monte Carlo samples \([5, 10, 50, 100]\), selecting \(\lambda = 0.1\) and \(M = 50\) based on the low validation loss.

\subsection*{Permuted Data}
We permute the data by first drawing a sequence of data, and then permuting it across the sequence. We tested data permutations of \([16, 32, 64]\) and chose 32, as it showed no significant difference and matched our batch size.

\end{document}